\documentclass[letterpaper]{article} 
\usepackage{aaai24}  
\usepackage{times}  
\usepackage{helvet}  
\usepackage{courier}  
\usepackage[hyphens]{url}  
\usepackage{graphicx} 
\usepackage{natbib}  
\usepackage{caption} 
\usepackage{algorithm}

\usepackage{newfloat}
\usepackage{listings}
\DeclareCaptionStyle{ruled}{labelfont=normalfont,labelsep=colon,strut=off} 
\floatstyle{ruled}
\newfloat{listing}{tb}{lst}{}
\floatname{listing}{Listing}
\usepackage{amssymb} 
\usepackage{amsmath}
\usepackage{mathtools}
\usepackage{tikz}
\usetikzlibrary{bayesnet}
\usepackage{mdframed}
\usepackage{xspace}
\usepackage{graphicx} 
\usepackage{pifont}
\usepackage{subcaption}
\usepackage{booktabs}
\usepackage{algpseudocode}
\usepackage{enumitem}
\usepackage{xcolor} 
\usepackage{svg}
\usepackage{arydshln}
\usepackage{multirow}
\usepackage{soul}

\newcommand{\approach}{{GO-DICE}\xspace}
\newcommand{\semiapproach}{{GO-DICE-Semi}\xspace}

\newcommand{\pnp}[1]{\texttt{PnP$\times${#1}}}
\newcommand{\stack}[1]{\texttt{Stack$\times${#1}}}

\definecolor{myred}{HTML}{DA4C4C}
\definecolor{mygreen}{HTML}{479A5F}
\definecolor{myyellow}{HTML}{EDB732}
\definecolor{myblack}{HTML}{000102}
\definecolor{mygrey}{HTML}{A1A9AD}
\definecolor{myblue}{HTML}{5BC5DB}

\title{GO-DICE: Goal-Conditioned Option-Aware Offline Imitation Learning via Stationary Distribution Correction Estimation}

\author {
    Abhinav Jain, 
    Vaibhav Unhelkar 
}
\affiliations {
    Rice University, Houston, TX\\
    abhinav.jain@rice.edu, vaibhav.unhelkar@rice.edu
}

\begin{document}

\maketitle

\begin{abstract}
Offline imitation learning (IL) refers to learning expert behavior solely from demonstrations, without any additional interaction with the environment.
Despite significant advances in offline IL, existing techniques find it challenging to learn policies for long-horizon tasks and require significant re-training when task specifications change.
Towards addressing these limitations, we present \approach an offline IL technique for goal-conditioned long-horizon sequential tasks.
\approach discerns a hierarchy of sub-tasks from demonstrations and uses these to learn separate policies for sub-task transitions and action execution, respectively; this hierarchical policy learning facilitates long-horizon reasoning.
Inspired by the expansive DICE-family of techniques, policy learning at both the levels transpires within the space of stationary distributions.
Further, both policies are learnt with goal conditioning to minimize need for retraining when task goals change.
Experimental results substantiate that \approach outperforms recent baselines, as evidenced by a marked improvement in the completion rate of increasingly challenging pick-and-place Mujoco robotic tasks.
\approach is also capable of leveraging imperfect demonstration and partial task segmentation when available, both of which boost task performance relative to learning from expert demonstrations alone.
\end{abstract}
\section{Introduction}

Learning to make decisions and accomplish sequential tasks is a core problem in artificial intelligence (AI).
Reinforcement learning (RL) addresses this problem by enabling agents to learn task policies by interacting with their environment.
However, in many practical scenarios, exploratory interaction with the environment is expensive, unsafe, or even infeasible.
For these scenarios, offline imitation learning (IL) offers AI agents an approach to learn task policies without any environmental interaction when task demonstrations provided by other (typically expert) agents are available.
Also referred to as learning from demonstration, classic techniques for offline IL include behavioral cloning (BC) and inverse reinforcement learning (IRL) \cite{pomerleau1991efficient, abbeel2004apprenticeship}.
These techniques and their extensions have been used to address complex problems in robotics, human modeling, and beyond \cite{unhelkar2020semi, ravichandar2020recent, wu2021learning}.

\begin{figure*}
\centering
    \includegraphics[width=0.99\textwidth]{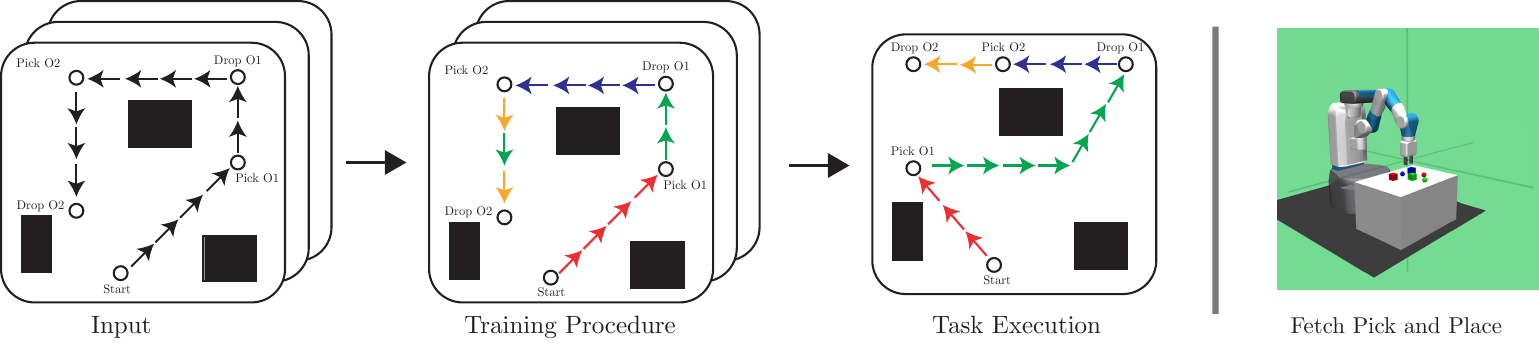}
\caption{(Left) Schematic illustration of \approach on a 2-dimensional 2-object pick and place task.
Given expert and imperfect demonstrations, iteratively, \approach segments them into sub-tasks (shown as colored arrows in the training procedure) and uses these segments to learn a goal-conditioned hierarchical policy. The task segmentation may be imperfect and improves as the policy estimate improves.
The learned policy can be used to execute tasks, even when the underlying goals (e.g., pick and place locations) change. (Right) A snapshot of the high-dimensional Fetch Pick-and-Place environment used in experiments.}
\label{fig:schematic} 
\end{figure*}

Originating almost three decades ago, offline IL remains an active area of research with recent techniques aiming to solve increasingly challenging sequential tasks \cite{osa2018algorithmic, arora2021survey, seo2022semi}.\footnote{Closely related is the paradigm of online imitation learning that combines demonstration data with agent's experience. However, like reinforcement learning, it is not suitable when interacting with environment is unsafe, expensive or infeasible.}
For instance, to enhance scalability and address tasks with high-dimensional state spaces, more recent IL techniques leverage advances in deep learning and generative adversarial training to represent and learn policies \cite{ho2016generative}.
Similarly, to address poor generalizability of classical techniques to the states absent in expert demonstrations \cite{ross2011reduction}, recent IL techniques have employed imperfect demonstrations to obtain a more comprehensive coverage of the state space in the training data \citep{kim2021demodice, kim2022lobsdice, ma2022versatile}.
Despite these advancements, existing offline IL techniques find it challenging to learn policies for long-horizon tasks and require significant retraining when task specifications change.

Towards addressing these limitations, we introduce a goal-conditioned option-aware approach to offline IL: \approach.
Inspired by the options framework~\cite{sutton1999between}, \approach segments available demonstrations into a sequence of sub-tasks to facilitate long-horizon reasoning.
It then uses the segmentation results to learn a hierarchy of policies for transitioning between sub-tasks and action execution within a sub-task.
The task segmentation and hierarchical policy learning repeats iteratively until convergence.
In \approach, policies at both the levels of hierarchy depend on not only state but also goals. 
Similar to recent RL techniques such as HER \cite{andrychowicz2017hindsight, fang2018dher}, this goal conditioning seeks to minimize the need for retraining when task goals change.

Given the task segmentations and choice of policy representation, a policy learning subroutine is necessary to learn each level of hierarchical policy.
\approach leverages a technique called DemoDICE for this purpose, owing to its ability to learn from imperfect demonstrations and without adversarial training \cite{kim2021demodice}.
As a member of the \textit{stationary DIstribution Correction Estimation} (DICE) family of approaches, \approach too learns within the space of stationary distributions.
However, in contrast to prior DICE-based techniques, \approach learns option-aware goal-conditioned policies and (when available) is able to utilize annotations of task segments to boost its learning.
In summary, we make three key contributions in the paper.
\begin{itemize}
    \item First, we provide an algorithm called \approach for offline imitation learning from (expert and imperfect) demonstrations and (when available) partial annotations of task segments. Fig.~\ref{fig:schematic} provides a schematic illustration.
    \item Second, through numerical experiments, we show that \approach outperforms recent baselines in challenging long-horizon Mujoco tasks (shown in Fig.~\ref{fig:schematic} right) and does not require retraining when task goals change.
    \item Third, through an ablation study, we show that \approach can successfully utilize partial annotations of task segments when available. In tasks where humans can provide these annotations, this feature of our approach can be used to boost learning performance.
\end{itemize}

\section{Related Work}
Before describing \approach, we briefly highlight the concepts and IL techniques that have informed our work.
A comparative overview is provided in Table \ref{tab:related-work}.

\paragraph{IL via stationary \textit{DI}stribution \textit{C}orrection \textit{E}stimation.}
The problem of IL has been mapped to a variety of optimization formulations.
For instance, a fruitful line of research has been the use of generative adversarial optimization, resulting in techniques such as GAIL and its extensions~\cite{ho2016generative}.
However, these techniques tend to require a large amount of training data, making them appropriate in settings where the learner can interact with its environment to collect this data.
For offline IL, the focus of this work, the DICE family of algorithms has recently gained momentum, which involve estimating the corrections between the stationary distribution of the optimal policy and that of the provided dataset~\cite{kostrikov2019imitation}.
For instance, DemoDICE and LobsDICE optimize KL-divergence regularized state-action and state-transition stationary distribution matching objectives, respectively~\cite{kim2021demodice, kim2022lobsdice}. 
SMODICE optimizes a more general f-divergence regularized state-occupancy only matching objective~\cite{ma2022versatile}.
In this work, we build on these DICE-family of techniques, owing to their state-of-the-art performance in offline IL and (relative to generative adversarial training) stable training process.

\begin{table}
\centering
\caption{An overview of related IL techniques}
\label{tab:related-work}
\resizebox{\columnwidth}{!}{%
\begin{tabular}{@{}lcccc@{}} \toprule
                         & \multicolumn{2}{c}{Policy Representation} & \multicolumn{2}{c}{Can Learn Using} \\ \midrule
                         & Goal-      & Option-      & Imperfect  & No Inter-\\ 
                         & conditioned & aware      & Demos.  & action \\ \midrule
GAIL

&                       &                   &                           &                \\
GoalGAIL                & $\checkmark$          &                   &                           &                \\
OptionGAIL              &                       & $\checkmark$      &                           &                \\
DemoDICE                 &                       &                   & $\checkmark$              & $\checkmark$   \\
GoFAR                    & $\checkmark$          &                   & $\checkmark$              & $\checkmark$   \\
\approach & $\checkmark$          & $\checkmark$      & $\checkmark$              & $\checkmark$  \\ \bottomrule
\end{tabular}%
}
\end{table}

\paragraph{Auxiliary Inputs in IL.}
Classically, IL aims to learn policies to solve Markov decision processes (MDPs) given \textit{expert} demonstrations~\cite{puterman2014markov, osa2018algorithmic}.
Mathematically, policies are represented as a mapping from states to actions.
Demonstrations correspond to (state, action)-tuples. 
More recently, guided by both computational and human-centered aspects, this classical paradigm has been extended to consider alternate policy representations and training data that involve auxiliary inputs.
For instance, imperfect demonstrations have been introduced as additional inputs to facilitate generalization~\cite{wu2019imitation, wang2021learning, kim2021demodice}.
Recognizing that human teachers can provide auxiliary inputs (such as corrections) easily in certain tasks, human-guided IL approaches have been developed~\cite{chernova2014robot,unhelkar2019learning,quintero2022human,habibian2022here}.
Our work is informed by such techniques and considers four types of auxiliary inputs -- goals, options, task segments, and imperfect demonstrations -- to facilitate generalization and long-horizon reasoning.

\paragraph{Goal-conditioned IL.}
Goal-conditioned policy representations enable learning of a unified policy for a family of tasks, which are identical in all respects but have distinct goals.
Goal-conditioned IL techniques have been developed using both GAIL and DICE frameworks.
GoalGAIL performs goal-conditioned occupancy measurement matching to learn a unified policy for a variety of task goals~\cite{ding2019goal}; similar to GAIL, this technique requires the learner to interact with its environment.
Among the DICE-family, the algorithm GoFAR performs goal-conditioned offline IL~\cite{ma2022offline}. 
In contast, the proposed \approach considers not only goals but also sub-tasks in its policy representation.

\paragraph{Option-aware IL.}
Long-horizon tasks can be viewed as composed of a sequence of sub-tasks that need to be executed in a particular order~\cite{byrne1998learning}.
This modeling insight has been formalized using the options framework~\cite{sutton1999between, daniel2016probabilistic} and offers an effective strategy to learning tasks by breaking them down into manageable sub-tasks, learning to accomplish them individually, and subsequently combining them to achieve the overarching task objective.
While initially proposed for RL, this hierarchical approach has also been utilized for IL~\cite{ranchod2015nonparametric, le2018hierarchical, unhelkar2019learning, gupta2019relay, jing2021adversarial, orlov2022factorial, jamgochian2023shail, nasiriany2023learning, gao2023transferring, chen2023option}.
Unsupervised methodologies, such as InfoGAIL and Directed InfoGAIL, leverage information-theoretic measures to initially discover latent options and subsequently imitate the expert~\cite{li2017infogail, sharma2018directed}.
OptionGAIL, a variant of note uncovers options concurrently through an Expectation-Maximization procedure~\cite{jing2021adversarial}.

Although this line of work has proven successful in settings where the learner can interact with the environment and has a fixed set of task goals, its efficacy in entirely offline IL scenarios with dynamically varying goals remains an open question that we explore in this work.
To our knowledge, \approach is the first approach to offline imitation learning that considers both goal-conditioned and option-aware policies.
Further, informed by practical considerations, \approach includes mechanisms to boost learning from auxiliary information (namely, annotations of task segments and imperfect demonstrations) when available.

\section{Problem Statement}

\paragraph{Task Model.}
We model the tasks of interest as goal-conditioned MDPs~\cite{schaul2015universal}.
Formally, the tasks are specified via the tuple $\big(\mathcal{S}, \mathcal{G}, \mathcal{A}, \mathbf{R}, \mathbf{T},\mu_{s}, \mu_g, \gamma, T \big)$, where $(\mathcal{S}, \mathcal{A}, \mathcal{G})$ are the set of task states, actions, and goals; $\mathbf{R}: \mathcal{S} \times \mathcal{A} \times \mathcal{G}\mapsto \mathbb{R}$ is the goal-conditioned reward; $\mathbf{T}: \mathcal{S}\times\mathcal{A}\mapsto\Delta(\mathcal{S})$ represents the transition function; $\mu_{s}(s)$ represents the initial state distribution; $\mu_g$ is the set of goals; $T$ task-horizon and $\gamma \in (0, 1]$ is the discount factor.
Typically, goal-conditioned MDPs utilize a sparse reward function, which is $1$ when goal is achieved and $0$ otherwise.

\paragraph{Expert Policy.}
Informed by the options framework, we assume that the expert solves this goal-conditioned task by completing a set of sub-tasks (or, equivalently, options).
Mathematically, in our model, the expert maintains 
\begin{itemize}
    \item a set of $K$ discrete options,  $c \in \mathcal{C}=\{1, \ldots, K\}$,
    \item initial option distribution, $c_0 \sim \mu_c(.|s, g)$,
    \item a high-level goal-conditioned policy for selecting the next option (or sub-task), $\pi_H(c|s,c',g)$, and 
    \item a set of $K$ low-level goal-conditioned policies for executing the chosen sub-task, $\pi_L(a|s,c,g)$.
\end{itemize}
We denote $\pi_E \equiv \{\pi_H, \pi_L\}$ to represent the expert's policy.

\paragraph{Inputs.}
The problem of offline IL assumes as inputs the task model without the reward function $\big(\mathcal{S}, \mathcal{G}, \mathcal{A}, \mathbf{T},\mu_s, \gamma \big)$ and a set of expert demonstrations, $\mathcal{D}_E$.
Each demonstration represents an execution trace, $\tau \in \mathcal{D}_E = \{s_{0:T}, a_{0:T-1}\}$ generated by the expert using $\pi_E$.
Besides expert demonstrations, we consider the following auxiliary inputs: $D_I$, a set of imperfect demonstrations collected with unknown degree of optimality; $g$, task goal for each demonstration $\tau \in D_I \cup D_E$; (optionally) the number of options $K$; and (optionally) partial annotations of task segments.%
\footnote{A \textit{task segment} is defined as a continuous sequence of state-action tuples associated with the same option. Specifically, a continuous sequence $(s_{t_1:t_2}, a_{t_1:t_2})$ is referred to as a task segment if (a) for all \( t \) in the interval \([t_1, t_2]\), \( c_{t} \) remains constant at some option \( c \in \mathcal{C} \) and (b) \( c_{t_1-1} \) and \( c_{t_2+1} \) are distinct from \( c \).}
Mathematically, partial annotations of task segments correspond to option labels $(c_{0:T-1})$ for the expert demonstrations $\tau \in \mathcal{D}_E$.

We refer to the problem as ``semi-supervised'' when the optional inputs are available.
The use of optional inputs is informed by prior human-guided IL techniques that leverage auxiliary inputs which can be readily queried from a human expert~\cite{chernova2014robot, unhelkar2019learning}.
In the option-aware setting, we observe that sub-tasks change less frequently over the course of a long-horizon demonstration; i.e., the number of change-points of options is significantly lower than the demonstration length.
Thus, in domains where a human expert can label change-points of sub-tasks, the optional input can be obtained with low annotation effort and help boost learning performance.
As such, semi-supervised techniques that can leverage this auxiliary information when available are desirable.

\paragraph{Desired Output.} Given the problem inputs, we consider the problem of learning an estimate of the expert policy $\pi_E$ without any interaction with the environment.

\section{GO-DICE: Goal-conditioned Option-aware stationary DIstribution Correction Estimation}

To solve this problem, we introduce the algorithm \textit{Goal-Conditioned Option-aware stationary DIstribution Correction Estimation} (\approach).
\approach extends the DICE family of offline IL techniques by incorporating goal conditioning and options.
To derive \approach, we begin with the formulation of its underlying optimization problem and then describe the computational approach to solve it.

\subsection{Optimization Problem}
Inspired by DICE family of techniques~\cite{kostrikov2019imitation}, \approach utilizes distribution matching to estimate the expert policy.
Briefly, DICE techniques seek to align the stationary distribution induced by the learned policy $d^\pi(\cdot)$ with that induced by the expert $d^{\pi_E}(\cdot)$.
Mathematically, this is achieved by minimizing the KL divergence between the two distributions.
As detailed in Related Work, a variety of stationary distributions have been previously considered, such as state occupancy $d^\pi(s)$, state-action occupancy $d^\pi(s,a)$, and state transitions $d^\pi(s,s')$.
We consider a stationary distribution that is goal-conditioned and option-aware,  $d^{\pi}(c_{-1},s,a,c; g)$
\begin{align*}
=(1-\gamma)\sum_{t=0}^{\infty}\gamma^tp\Big(c_{t-1}=c_{-1}, s_t=s, a_t=a, c_t=c; g). 
\end{align*}
Three stationary distributions are of interest: namely, $d^{\pi_E}$, that induced by the expert policy; $d^{\pi_O}$ that induced by the data set consisting of expert and imperfect demonstrations $\mathcal{D}_O = \mathcal{D}_E \cup \mathcal{D}_I$; and that induced by the learned policy $d^{\pi}$.

\paragraph{Primal Optimization Problem.}
Given the stationary distributions, we can now define the optimization problem for \approach.
Informed by work of \cite{kim2021demodice}, which also considers DICE-based learning from imperfect demonstrations albeit without goals or options, \approach seeks to solve the following constrained optimization problem
\begin{align}
     \min_{d^{{\pi}}} D_{KL}(d^{{\pi}}||d^{\pi_E})  + \alpha D_{KL}(d^{{\pi}}||d^{\pi_O})
\label{eq:il_dist_match_reg}
\end{align}
subject to Bellman constraints: $\sum_{c, a} d^{\pi} (c', s, c, a; g) = (1$
$- \gamma){\mu}(c', s; g) + \gamma\sum_{c'', s', a'} \mathbf{T}(s|s',a')d^{{\pi}}(c'', s', c', a'; g)$ and
$d^{{\pi}}(c', s, c, a; g) \geq 0 \; \forall \; c, c'\in\mathcal{C}, s\in \mathcal{S}, a\in\mathcal{A}, g \in \mathcal{G}$.
Intuitively, the optimization function seeks to minimize the difference between stationary distributions induced by the expert and learnt policy.
The imperfect demonstrations are used as a regularizer through the term $\alpha D_{KL}(d^{{\pi}}||d^{\pi_O})$, where $\alpha$ is the regularization coefficient.
This distribution regularization has been shown to facilitate offline IL by penalizing distribution drift, without the need of any on-policy sampling~\citep{nachum2019algaedice, lee2021optidice, kim2021demodice, kim2022lobsdice, ma2022offline, ma2022versatile}.
The Bellman constraints ensure that $d^{{\pi}}$ is a valid stationary distribution induced by an agent following the hierarchical policy ${\pi}$.

\paragraph{Primal to Dual Conversion.}
For tractable optimization, we next convert the constrained optimization problem of Eq.~\ref{eq:il_dist_match_reg} into its dual unconstrained formulation.
Using the method of Lagrange multipliers, we obtain:
\begin{align}
     &\max_{d^{{\pi}}\geq0}\min_{\nu}f(\nu, d) {-} D_{KL}(d^{{\pi}}||d^{\pi_E}) {-} \alpha D_{KL}(d^{{\pi}}||d^{\pi_O}) \label{eq:lagrangian function} \\
     &\text{where} \; f(\nu, d^{{\pi}}) = \sum_{c',s,g}\nu(c',s,g)\bigg((1-\gamma){\mu}(c', s; g) +  \nonumber \\
     &\phantom{f(\nu,  d^{{\pi}}) = \sum_{c',s,g}} \gamma (T_*d^{{\pi}})(c',s, g) - \sum_{c, a} d^{\pi}(c', s, c, a; g)\bigg), \nonumber
\end{align}%
$\nu(c', s, g)$ are the Lagrangian multipliers; and $T_*$ is the transposed Bellman operator such that 
\begin{equation}
(T_*d)(c',s; g) = \sum_{c'', s', a'}\mathbf{T}(s|s',a')d(c'', s', c', a'; g).
\end{equation}

The dual objective can be further simplified by introducing the notation of stationary distribution ratio
\begin{align}
w(c',s,c,a, g) = \frac{d^{{\pi}}(c',s,c,a; g)}{d^{\pi_O}(c',s,c,a; g)}.\label{eq:imp_weight} 
\end{align}
Using stationary distribution ratio $w$, Eq.~\ref{eq:lagrangian function} translates to the following maximin optimization problem
\begin{align}
   \max_{w\geq 0}&\min_{\nu} \; (1-\gamma)\mathbb{E}_{{\mu}}[\nu(c',s; g)] + \mathbb{E}_{d^{\pi_O}}\Big[w(c',s,c,a, g) \nonumber \\
   &\big(A_{\nu}(c',s,c,a, g) - (1+\alpha)\log w(c',s,c,a, g)\big)\Big]
   \label{eq:new lagrangian function}
\end{align}
where, $A_\nu$ resembles the advantage function and is given as
\begin{align}
    A_{\nu} & =r(c',s,c,a, g){+}\gamma (T\nu)(s, c, a, g) {-} \nu(c',s, g) \label{eq:advantage_fn}\\
    r &= \log\frac{d^{\pi_E}(c',s,c,a; g)}{d^{\pi_O}(c',s,c,a; g)} \label{eq:log_distribution_ratio}\\   
    (T\nu)&(s, c, a;g)=\sum_{s'} \mathbf{T}(s'|s, a)\nu(c, s', g).
\end{align}

\paragraph{Conversion to Direct Convex Optimization.}
While the dual formulation avoids the need of constrained optimization, it still requires solution of a challenging maximin optimization problem.
Similar to \cite{kim2021demodice}, we seek to enhance the stability of the optimization process by reducing it to a direct convex optimization problem.
A detailed derivation of this conversion is provided in the Appendix, which results in the following direct convex optimization problem:
\begin{equation}
    \min_{\nu} \; (1-\gamma)\mathbb{E}_{{\mu}}[\nu(c',s, g)] {+} (1+\alpha)\mathbb{E}_{d^{\pi_O}}[w^*(\cdot)]
    \label{eq:value_func_objective_unstable}
\end{equation}
where, $w^*$ denotes the optimal importance weights given as:
\begin{align}
    w^*(c'{,}s{,}c{,}a{;}g) {=} \frac{d^{{\pi^*}}(c',s,c,a; g)}{d^{\pi_O}(c',s,c,a, g)} {=} \exp\bigg(\frac{A_{\nu}}{1{+}\alpha}{-}1\bigg)
   \label{eq:optimal_imp_weight}
\end{align}
This problem can be further stabilized via the following surrogate objective, which shares the same optimal value with Eq.~\ref{eq:value_func_objective_unstable} but is less prone to exploding gradients:
\begin{equation}
    \min_{\nu} \; (1-\gamma)\mathbb{E}_{{\mu}}[\nu(\cdot)] {+} (1+\alpha)\log\mathbb{E}_{d^{\pi_O}}[\exp(\frac{A_{\nu}}{1+\alpha})]
    \label{eq:value_func_objective}
\end{equation}

\subsection{Learning Algorithm}
Having defined the optimization problem, we now describe the computational approach to solve it.
Algorithm~\ref{alg:godice} provides an overview of the approach, which trains four classes of neural networks: $\pi, \pi', \Psi, \nu$ to estimate the expert policy.
Given the problem inputs, the algorithm first segments all available demonstrations into sub-tasks through a Viterbi-style subroutine (lines 4-7).
Using the results of the task segmentation and demonstrations, the algorithm then updates the discriminator network, Lagrange multipliers, and the policy networks via gradient descent (lines 8-10).
This process repeats iteratively until convergence.
In the remainder of this section, we detail each subroutine of the algorithm along with the associated hyperparameters.

\begin{algorithm}[t]
\caption{\approach}
\label{alg:godice}
\begin{algorithmic}[1]
\Require Expert trajectories $\mathcal{D}_E$, imperfect trajectories $\mathcal{D}_I$, task goals for each trajectory, and (optional) partial annotations of task segments
\Ensure Learned policy ${\pi} = \{\pi_H, \pi_L\}$
\State \textbf{Parameters:} $\alpha, \lambda, K, M, N$
\State \textbf{Initialize:} $\pi, \pi_t, \Psi, \nu$
\For{$n = 1$ to $N$}
    \If{every $M$ iterations}
        \State ${\pi}' \gets \lambda{\pi}' + (1-\lambda){\pi}$ \Comment{Update target networks}
        \State Segment unannotated trajectories using Eq.~\ref{eq:Viterbi}
    \EndIf
    \State Update discriminator network $\Psi$ using Eq.~\ref{eq:disc_func_objective}
    \State Update Lagrange multipliers $\nu$ using Eq.~\ref{eq:value_func_objective}
    \State Update policy networks ${\pi}$ using Eq.~\ref{eq:policy_func_objective}
\EndFor
\end{algorithmic}
\end{algorithm}

\paragraph{Estimating the Lagrange Multipliers.}
We represent the Lagrange multiplier $v(c',s,g)$ as a neural network.
The network parameters are updated by solving Eq.~\ref{eq:value_func_objective} via gradient descent.
Computing the associated gradients requires estimates of options $(c',c)$ and the advantage function $A_v$.
We defer the discussion of option inference to end of this section.
To compute the advantage function, we first train a discriminator network $\Psi (c', s, c, a;g) : \mathcal{C}\times\mathcal{S}\times\mathcal{C}\times\mathcal{A}\times\mathcal{G}\mapsto (0,1)$ with the following objective:
\begin{align}
    \min \quad &\mathbb{E}_{d^{\pi_E}}[\log\Psi(\cdot)]+ \mathbb{E}_{d^{\pi_O}}[\log(1-\Psi(\cdot))]
    \label{eq:disc_func_objective}
\end{align}
The optimal discriminator corresponds to $\Psi^*=\frac{d^{\pi_O}}{d^{\pi_O} + d^{\pi_E}}$, and thus can be used to estimate the log-distribution ratio defined in Eq.~\ref{eq:log_distribution_ratio} as $r = -\log\Big(\frac{1}{\Psi^*} - 1\Big)$.
Given $r$, we can compute the advantage function and required gradient using Eq.~\ref{eq:advantage_fn} and Eq.~\ref{eq:value_func_objective}, respectively.

\paragraph{Weighted Policy Learning.}
To update the policy networks, $\pi= \{\pi_H, \pi_L\}$, \approach performs weighted-behavior cloning using the following objective
\begin{align}
    \max\mathbb{E}_{d^{{\pi}^*}}&\big[\log{\pi}(c, a|s,c',g)\big] \nonumber\\
    = \max\mathbb{E}_{d^{{\pi}_O}}&\bigg[w^*(\cdot)\Big(\log\pi_H(\cdot) + \log\pi_L(\cdot)\Big)\bigg]
    \label{eq:policy_func_objective}
\end{align}
where $w^*$ denotes the optimal importance weights of Eq.~\ref{eq:optimal_imp_weight}.

\paragraph{Sub-task Inference.}
Subroutines for learning the various networks rely on not only demonstrations but also the labels of sub-tasks for these demonstrations.
Following \cite{jing2021adversarial}, we utilize Viterbi-style decoding to infer these labels
\begin{align}
    \hat{c}_{0:T-1} = \arg\max p(c_{0:T-1} | s_{0:T}, a_{0:T-1},  g)
\end{align}
for the unannotated trajectories. 
To solve this decoding problem, we define forward messages $\alpha_t(c_t)$ and utilize the following recursive formulation to compute them:
\begin{align} \label{eq:Viterbi}
    \alpha_t(c_t) &= \max_{c_{0:t-1}}\log p(c_t, a_{0:t}|s_{0:t}, c_{0:t-1}, g)\\
    & = \max_{c_{t-1}}\alpha_{t-1}(c_{t-1}) {+} \log\pi_H'(c_t|\cdot) {+} \log\pi_L'(a_t|\cdot) \nonumber
\label{eq:Viterbi}
\end{align}
where, $\alpha_0(c_0) = \log\mu_c'(c_0|s_0, g) +  \log\pi_L'(a_0|s_0, c_0, g)$.
Once all messages are computed, we employ a back-tracing method to decode the sequence of sub-tasks by maximizing $\alpha_t(c_t)$, starting from $\hat{c}_{T-1} = \arg\max_c \alpha_{T-1}(c)$.

Note that the decoding procedure requires an estimate of the expert policy.
Instead of using the latest policy estimate $\pi$, \approach utilizes target policy networks $\pi'$ in the option decoding procedure.
The target networks are synchronized every $M$ iterations with the main networks $\pi$ using $\lambda$-weighted Polyak averaging, where $M$ and $\lambda$ are hyperparameters (lines 4-5, Algorithm~\ref{alg:godice}).
Unlike other DICE techniques, the optimization routines of discriminator, Lagrange multipliers, and the policy in \approach are intertwined as they all depend on the decoded option labels.
Due to this iterative nature of \approach, we found target networks to be critical for enhancing the stability and convergence of the underlying optimization process.

\begin{figure*}[t]
\centering
\begin{subfigure}{.32\textwidth}
  \centering
    \caption{\pnp{1}}
  \includegraphics[width=\linewidth]{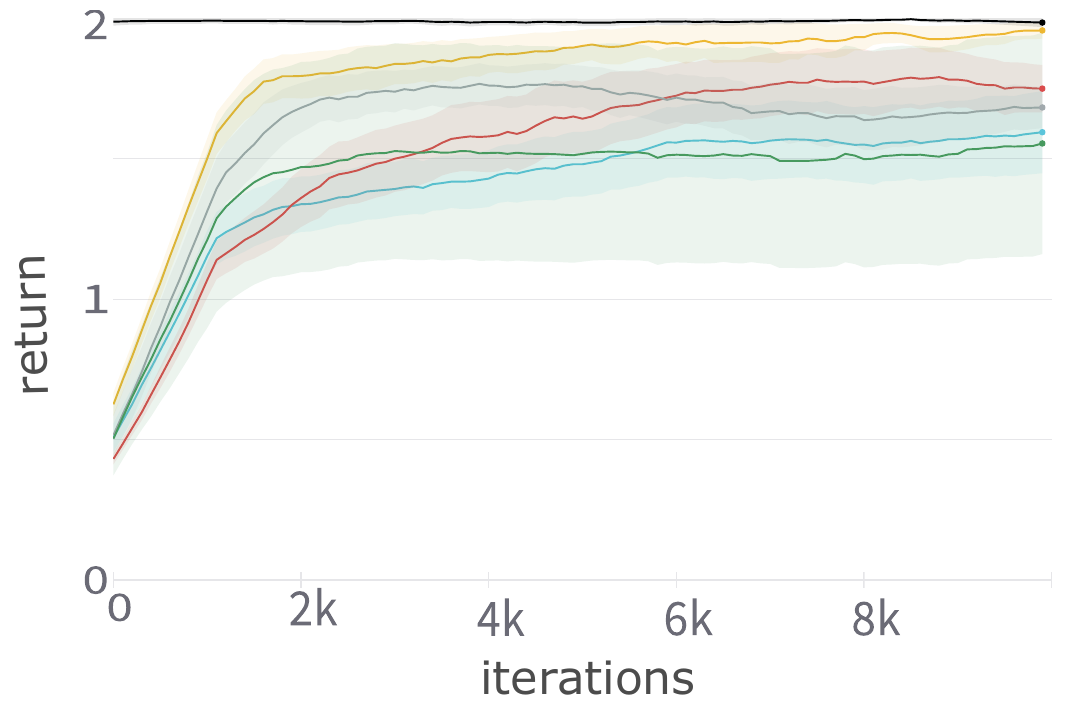}
  \label{subfig:PnP One Object}
\end{subfigure}%
\hfill
\begin{subfigure}{.32\textwidth}
  \centering
    \caption{\pnp{2}}
  \includegraphics[width=\linewidth]{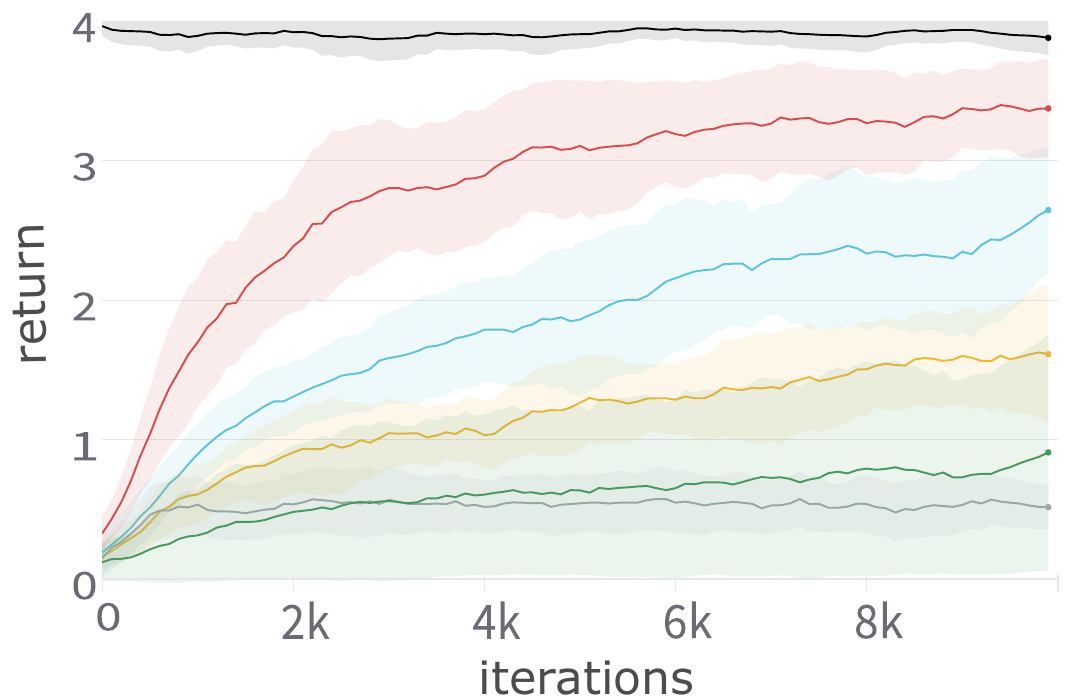}
  \label{subfig:PnP Two Object}
\end{subfigure}%
\hfill
\begin{subfigure}{.32\textwidth}
  \centering
    \caption{\pnp{3}}
  \includegraphics[width=\linewidth]{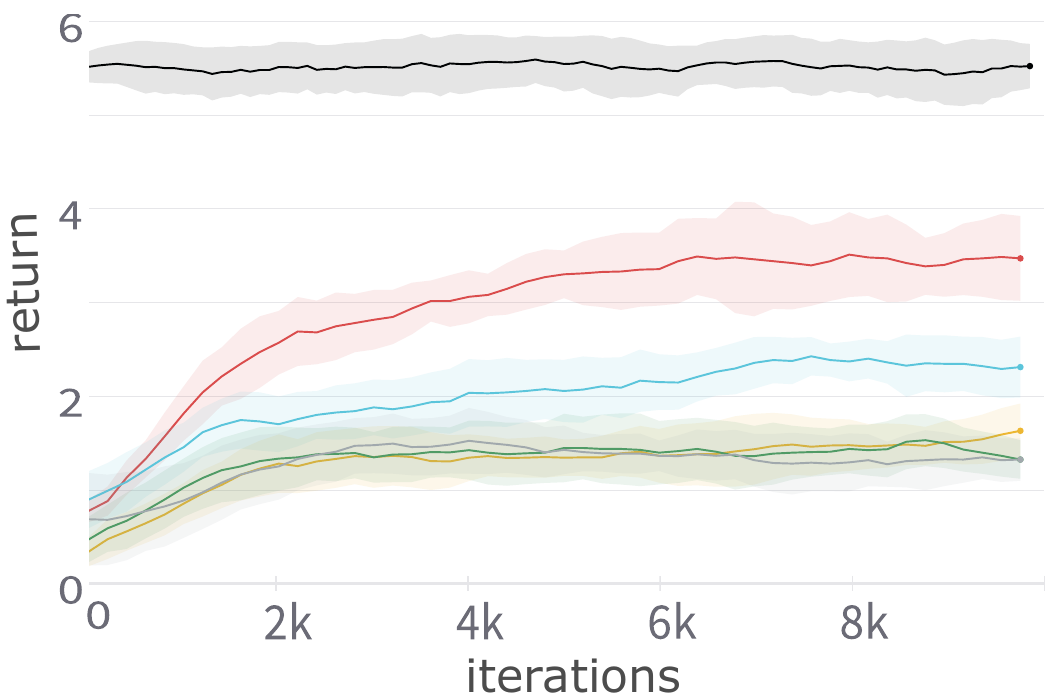}
  \label{subfig:PnP Three Object}
\end{subfigure}
\begin{tikzpicture}[overlay]
        
        \draw[myblack,line width=1pt] (-15.0cm,-0.cm) -- (-14.5cm,-0.cm);
        \node[anchor=west] at (-14.5cm,-0.cm) {\text{Expert}};

        \draw[mygrey,line width=1pt] (-13.0cm,-0.cm) -- (-12.5cm,-0.cm);
        \node[anchor=west] at (-12.5cm,-0.cm) {\text{BC}};

       \draw[myyellow,line width=1pt] (-11.5cm,-0.cm) -- (-11.0cm,-0.cm);
        \node[anchor=west] at (-11.0cm,-0.cm) {\text{g-DemoDICE}};

        \draw[mygreen,line width=1pt] (-8.5cm,-0.cm) -- (-8.0cm,-0.cm);
        \node[anchor=west] at (-8.0cm,-0.cm) {\text{GoFar}};

        \draw[myblue,line width=1pt] (-6.5cm,-0.cm) -- (-6.0cm,-0.cm);
        \node[anchor=west] at (-6.0cm,-0.cm) {\approach};

        \draw[myred,line width=1pt] (-4.0cm,-0.cm) -- (-3.5cm,-0.cm);
        \node[anchor=west] at (-3.5cm,-0.cm) {\approach\text{-Semi}};
\end{tikzpicture}
\caption{Comparative Performance: Learning curves of \approach and the baseline techniques for the three benchmark tasks. In all plots, the $x$-axis denotes the training iteration and $y$-axis denotes the return accrued by the learned policy.}
\label{fig:R1}
\end{figure*}
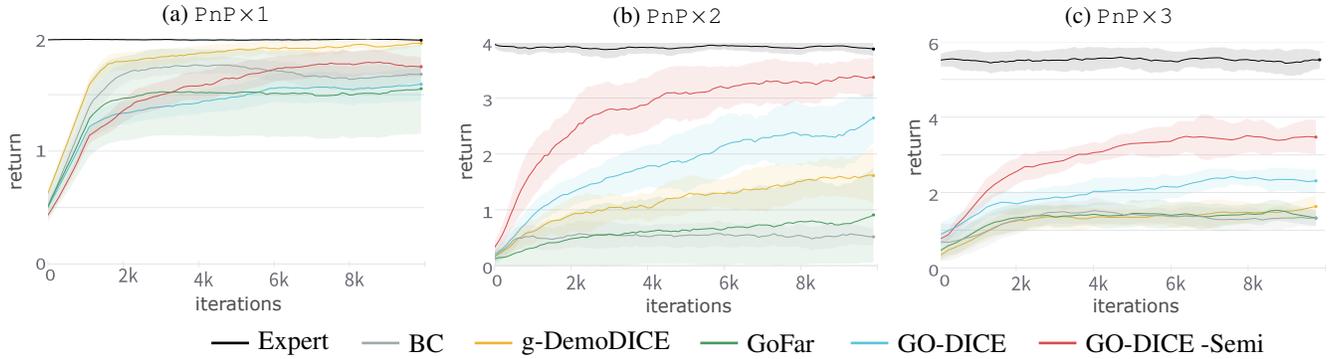

\section{Experiments}
We evaluate \approach on robotic manipulation tasks simulated using Mujoco~\cite{todorov2012mujoco}.%
\footnote{Please see the Appendix for additional details regarding experimental tasks, implementation details, and results.}
The tasks are modeled after the benchmark task Fetch Pick and Place (\texttt{PnP}), which require a robot to pick an object and place them in desired goal location~\cite{plappert2018multi}.
We consider three variants of this benchmark task -- \pnp{1}, \pnp{2}, \pnp{3} -- which include $1, 2,$ and $3$ objects, respectively.
The task complexity increases with the number of objects, requiring increasing levels of long-horizon reasoning.
Further adding to the complexity, desired goals (place locations of objects) changes across demonstrations.
These tasks are chosen for evaluation due to the fact that even the simplest variant \pnp{1} is challenging for offline IL algorithms~\cite{ding2019goal}.
Moreover, \texttt{PnP} family of tasks naturally encapsulate other primitive tasks commonly used in IL benchmarking, such as reach and grasp; thus, success in \texttt{PnP} requires the learner to also succeed in these primitive tasks.
Finally, by varying the number of objects, these tasks allows us to isolate the challenge of long horizon.

\paragraph{Training Data.}
Offline IL requires a set of demonstrations as training input.
To arrive at this data, we first create a hand-crafted expert policy for each task.
Expert demonstrations $\tau \in \mathcal{D}_E$ are generated using this expert policy, ensuring successful task completion.
Imperfect demonstrations $\tau \in \mathcal{D}_I$ are generated via two sources: noisy version of expert policy and randomly generated policies.
For \pnp{1} and \pnp{2}, the data includes $25$ expert and $75$ imperfect demonstrations.
For the more challenging \pnp{3}, the data includes 50 expert and 100 imperfect demonstrations.

\paragraph{Baselines.}
We benchmark against three offline imitation learning techniques: Behavioral cloning \text{(BC)}, \text{GoFAR}, and \text{g-DemoDICE}.
\text{GoFAR} is the most recent IL technique in the DICE family and learns goal-conditioned policies~\cite{ma2022offline}.
\text{g-DemoDICE} is a one-option equivalent of our algorithm.
It can be seen as a simple extension of \text{DemoDICE}~\cite{kim2021demodice} by incorporating goal conditioning.
Similar to \text{DemoDICE}, it is designed to learn from both expert and imperfect demonstrations.
All baselines and our technique receive the goal-conditioned demonstrations (i.e., $\mathcal{D}_I, \mathcal{D}_E$ with $g$) as inputs.
To evaluate the ability to learn from auxiliary inputs, we also compare against the semi-supervised version of our approach, denoted as \semiapproach.
This version also uses Algorithm~\ref{alg:godice} but additionally receives number of options $K_{\text{\semiapproach}}$ and sub-task labels for expert demonstrations (but not the imperfect demonstrations) as inputs.
$K_{\text{\semiapproach}}=3,6,$ and $9$ was provided as an input for the three tasks, respectively.

\paragraph{Hyper-parameters.}
Each model underwent training for $N=10k$ iterations.
All DICE-based algorithms were regularized with a replay value of $\alpha=0.05$.
For \approach, target network updates were managed using $(M, \lambda)$ pairs: $(20, 0.95)$ for one-object, $(20, 0.5)$ for two-object, and $(50, 0.5)$ for three-object PnP tasks.
The optimal values of the $(M, \lambda)$-tuple were chosen through a grid search of hyperparameters.
After parameter tuning, the option counts in \approach were selected as $K=2$, $3$, and $9$ for the one-, two-, and three-object tasks, respectively.

\paragraph{Evaluation Metric.}
Performance is quantified via cumulative reward (averaged over 10 evaluation episodes), where each successful pick or place earns a reward of $1$.
The reward function is not known to the learning algorithms.

\paragraph{Expert Policy.}
Guided by the experimental evaluations of \cite{ding2019goal}, we train an expert policy to produce expert demonstrations.
This trained expert is able to achieve near-optimal performance.
The variations in expert performance (Fig. \ref{fig:R1}) arise due to the sub-optimality of the expert policy in situations where the target goals of objects are in close proximity.
\section{Results}
We now pose several research questions, conduct experiments to answer them, and finally present our findings.

\paragraph{R1. How does the performance of \approach measure against the baselines in long-horizon tasks?}
\label{sec:R1}
Fig. \ref{fig:R1} reveals a distinct trend.
Most techniques perform near-optimally in the single-object task as seen in Figs.~\ref{subfig:PnP One Object}.
However, the performance disparity between \approach and its baselines amplifies with the introduction of multi-object tasks, evident from Figs.~\ref{subfig:PnP Two Object} and \ref{subfig:PnP Three Object}.
The driving force behind this difference can be linked to \approach's adeptness in discovering and decoding sub-tasks (such as reaching, grabbing, and placing objects) and seamlessly transitioning between them.
However, the performance of \approach diminishes in the three-object tasks.
One reason for this decrement could be the close object proximities leading to unintended collisions, displacing already placed objects. This behavior can potentially be rectified with self-correcting expert demonstrations, an aspect we plan to explore in the future.
When equipped with expert-driven trajectory segmentation, \semiapproach realizes even greater efficiency as reflected in its elevated average returns.
\textit{In conclusion, this experiment suggests that \approach is capable of discerning and executing sub-task hierarchies, which are pivotal for successfully tackling long-horizon tasks.}

\begin{figure}[t]
\begin{subfigure}{0.7\linewidth}
  \includegraphics[height=0.7\linewidth]{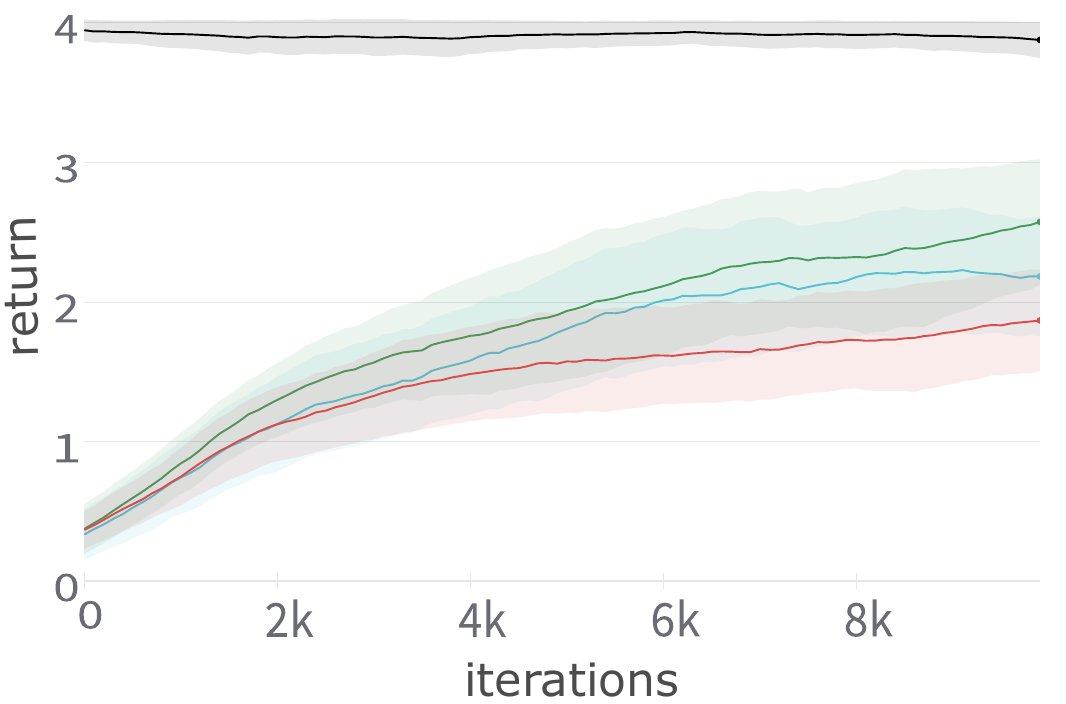}
  \caption{\pnp{2}}
  \label{subfig:R2_twoObj}
\end{subfigure} 
\begin{subfigure}{0.7\linewidth}
  \includegraphics[height=0.7\linewidth]{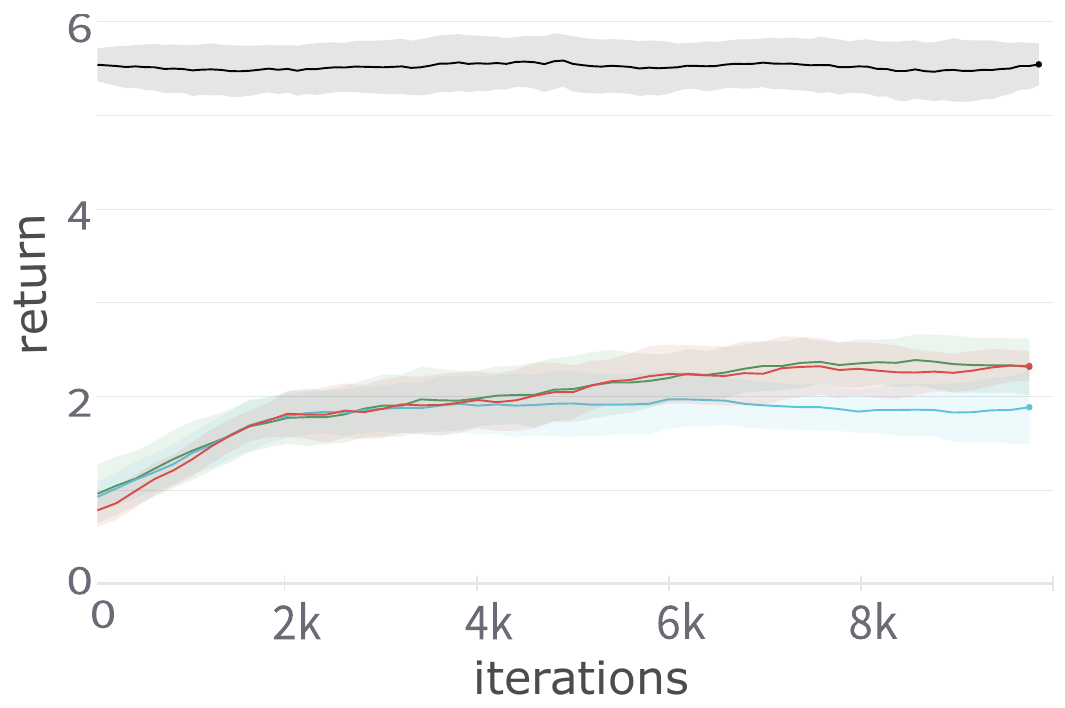}
  \caption{\pnp{3}}
  \label{subfig:R3_threeObj}
\end{subfigure}%
\begin{tikzpicture}[overlay]
        \node[anchor=west] at (0.5cm,8.2cm) {\pnp{2}};
        
        \draw[myblue,line width=1pt] (0.7cm,7.4cm) -- (1.2cm,7.4cm);
        \node[anchor=west] at (1.4cm,7.4cm) {$K=2$};

        \draw[mygreen,line width=1pt] (0.7cm,6.7cm) -- (1.2cm,6.7cm);
        \node[anchor=west] at (1.4cm,6.7cm) {$K=3$};

        \draw[myred,line width=1pt] (0.7cm,6.0cm) -- (1.2cm,6.0cm);
        \node[anchor=west] at (1.4cm,6.0cm) {$K=6$};

        \node[anchor=west] at (0.5cm,3.7cm) {\pnp{3}};
        
        \draw[myblue,line width=1pt] (0.7cm,2.9cm) -- (1.2cm,2.9cm);
        \node[anchor=west] at (1.4cm,2.9cm) {$K=3$};

        \draw[mygreen,line width=1pt] (0.7cm,2.2cm) -- (1.2cm,2.2cm);
        \node[anchor=west] at (1.4cm,2.2cm) {$K=6$};

        \draw[myred,line width=1pt] (0.7cm,1.5cm) -- (1.2cm,1.5cm);
        \node[anchor=west] at (1.4cm,1.5cm) {$K=9$};
\end{tikzpicture}
\caption{\approach: Effect of $K$ on the return of learned policy ($y$-axis). The $x$-axis denotes the training iteration.}
\label{fig:R2}
\end{figure}
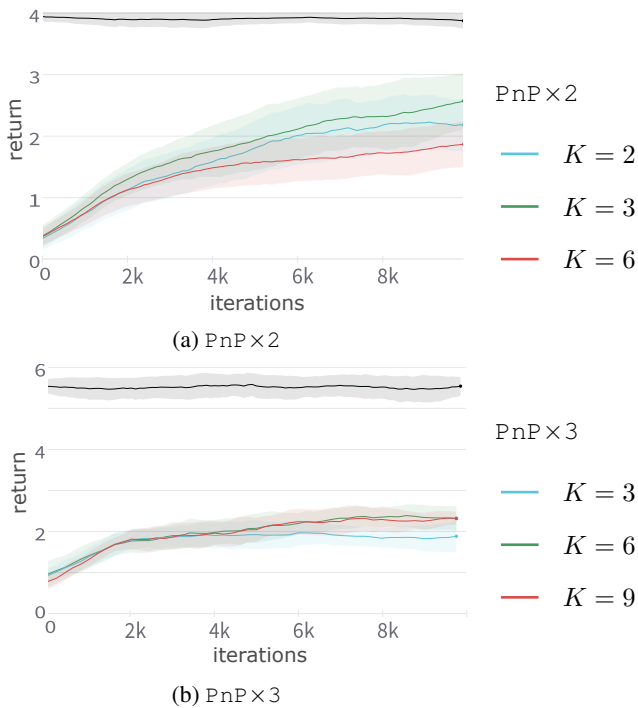
\paragraph{R2. How does the choice of hyperparameter $K$ affect \approach's learning performance?}
\label{sec:R2}

\approach differentiates from baselines primarily by leveraging a set of $K$ discrete options for task segmentation.
In general, $K$ is unknown that needs to be set as a hyperparameter.
Through this experiment, we investigate the effect of $K$ on model's convergence and overall performance, understanding if a specific option count optimally balances complexity with efficiency.
In particular, we train \approach with varying number of options.
From Figs.~\ref{subfig:R2_twoObj}-\ref{subfig:R3_threeObj} it is evident that merely augmenting the number of options does not guarantee improved performance. 
This is demonstrated by the suboptimal results with $K = 6$ options in \pnp{2} and no improvement for $K = 9$ options in \pnp{3}.
Yet, when expert segmentations for these options are introduced in a semi-supervised setting, there is a marked enhancement in performance, as seen in Figs. \ref{subfig:PnP Two Object}-\ref{subfig:PnP Three Object}.
\textit{These results suggest that increasing the option count with no supervision can boost performance up to a certain threshold.
Beyond this, performance may decline, likely because redundant transitions overshadow the benefits of expressive task segmentation.}

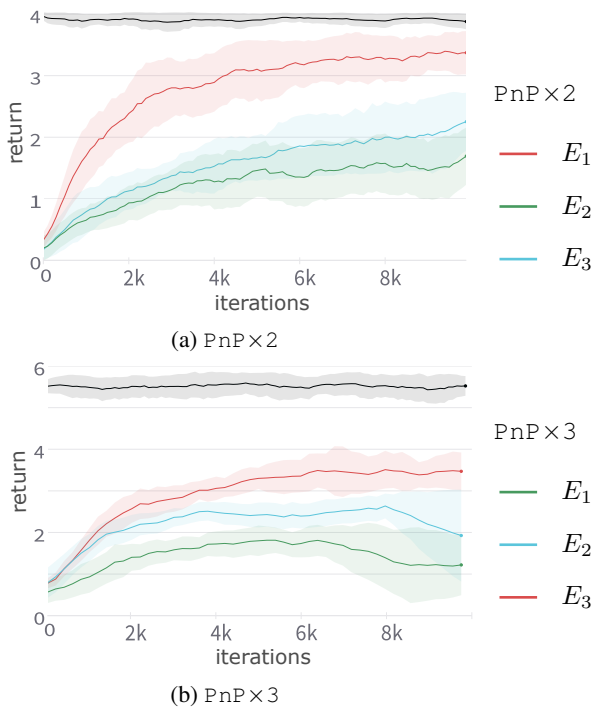
\begin{figure}[t]
\begin{subfigure}{0.7\linewidth}
  \includegraphics[height=0.7\linewidth]{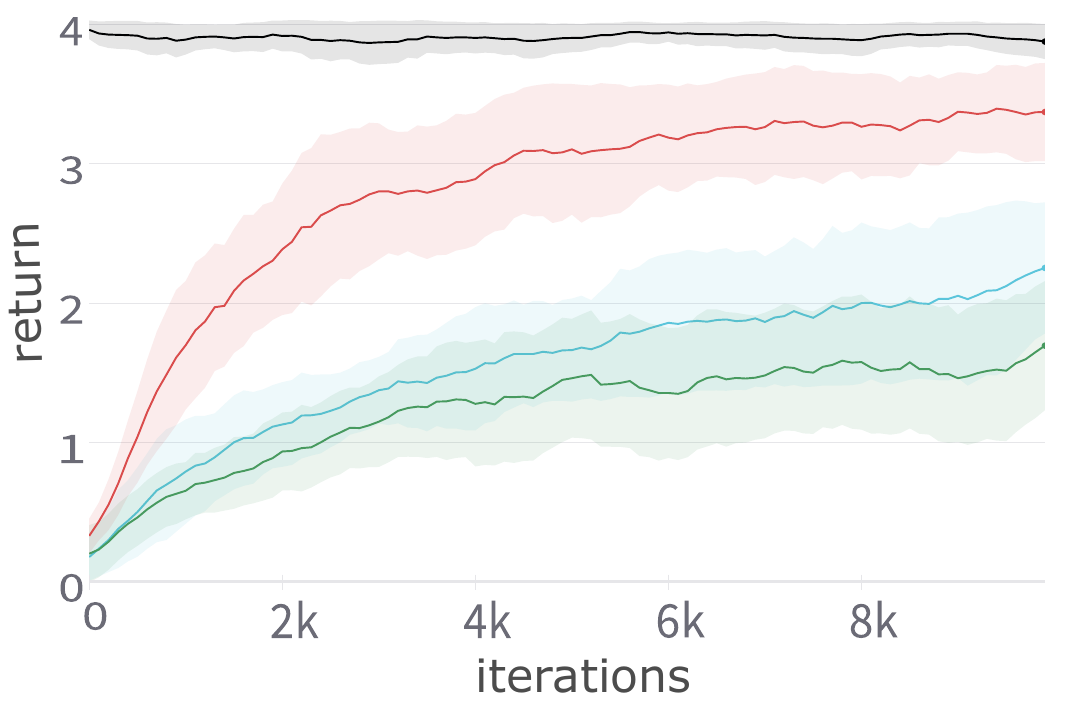}
  \caption{\pnp{2}}
\end{subfigure}
\begin{subfigure}{0.7\linewidth}
  \includegraphics[height=0.7\linewidth]{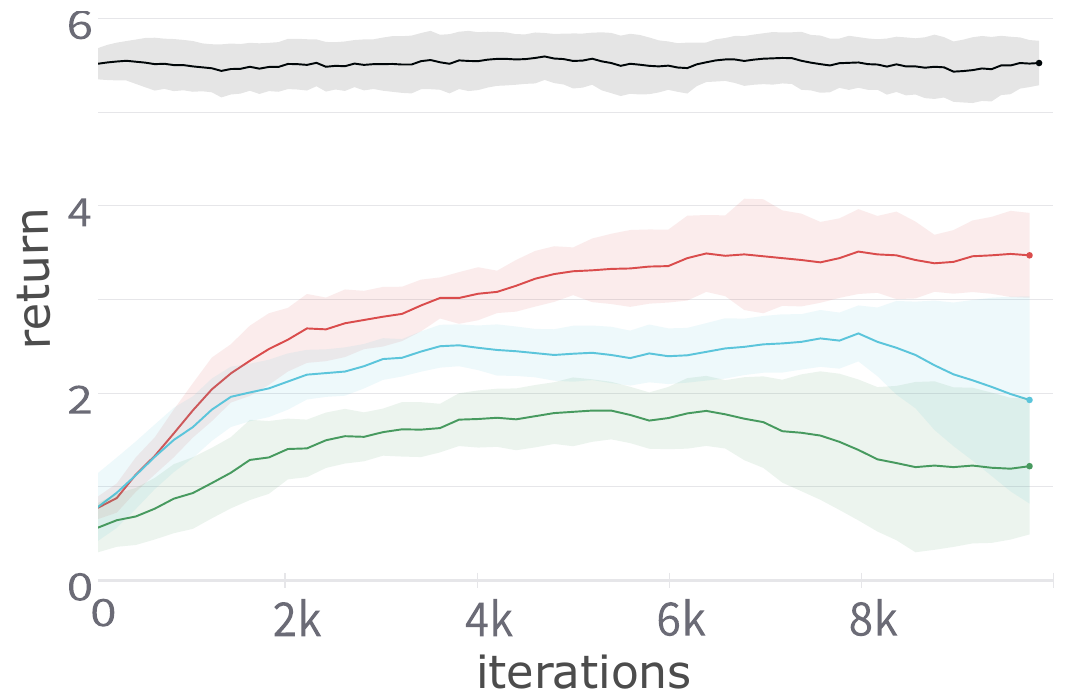}
  \caption{\pnp{3}}
\end{subfigure}%
\begin{tikzpicture}[overlay]
        \node[anchor=west] at (0.5cm,8.2cm) {\pnp{2}};

        \draw[myred,line width=1pt] (0.7cm,7.4cm) -- (1.2cm,7.4cm);
        \node[anchor=west] at (1.4cm,7.4cm) {$E_1$};
        
        \draw[mygreen,line width=1pt] (0.7cm,6.7cm) -- (1.2cm,6.7cm);
        \node[anchor=west] at (1.4cm,6.7cm) {$E_2$};

        \draw[myblue,line width=1pt] (0.7cm,6.0cm) -- (1.2cm,6.0cm);
        \node[anchor=west] at (1.4cm,6.0cm) {$E_3$};

        \node[anchor=west] at (0.5cm,3.7cm) {\pnp{3}};
        
        \draw[mygreen,line width=1pt] (0.7cm,2.9cm) -- (1.2cm,2.9cm);
        \node[anchor=west] at (1.4cm,2.9cm) {$E_1$};

        \draw[myblue,line width=1pt] (0.7cm,2.2cm) -- (1.2cm,2.2cm);
        \node[anchor=west] at (1.4cm,2.2cm) {$E_2$};

        \draw[myred,line width=1pt] (0.7cm,1.5cm) -- (1.2cm,1.5cm);
        \node[anchor=west] at (1.4cm,1.5cm) {$E_3$};
\end{tikzpicture}
\caption{\semiapproach: Effect of sub-task labeling on return of learned policy (denoted on $y-$axis).}
\label{fig:R3}
\end{figure}

\paragraph{R3. How does sub-task labeling affect \semiapproach's learning performance?} \label{sec:R3}
The semi-supervised version of \approach boosts learning performance using sub-task labels provided by human experts.
The notion of sub-task may differ across annotators.
As such, we aim to identify and distinguish various means of segmentation and explore their influence on performance.
To conduct these experiment, we identify three core primitives for the pick and place tasks: \texttt{reach}, \texttt{grasp}, and \texttt{place}.
Based on these primitives, we explore three sub-task labeling methods: (i) based on the primitive alone: $\mathcal{C}^{E_1} = \{\texttt{reach}, \texttt{grasp}, \texttt{place}\}$; (ii) based on the object being manipulated: $\mathcal{C}^{E_2} = \{\texttt{object}_1, \ldots, \texttt{object}_n\}$; and (iii) based on both primitive and object:  $\mathcal{C}^{E_3} = \{\texttt{reach}_{\texttt{object}_i}, \texttt{grasp}_{\texttt{object}_i}, \texttt{place}_{\texttt{object}_i}\}_{i=1}^{n}$.
These three methods offer task segmentation at different levels of granularity and from different perspectives.
Fig.~\ref{fig:R3} demonstrates that an expert-curated, finely-segmented sub-task labels $(E_3)$ yield superior performance.
At the same time, Figure \ref{fig:R2} indicates that simply increasing $K$ is insufficient. 
\textit{This suggests expert guidance becomes particularly critical when tasks demand finer segmentations. Further, Figure \ref{fig:R3} also confirms that \semiapproach is able to learn from different types of sub-task annotations.}

\paragraph{R4. Zero-shot Transfer: Can low-level policies from simpler tasks be used to execute more complex tasks without additional training?}
In this final experiment, we assess whether low-level policies learnt using \approach are transferable to more complex tasks that utilize similar primitives.
This experiment is exploratory, as \approach is not designed for zero-shot transfer.
Towards this question, we first train policies for \pnp{1}, \pnp{2}, and \pnp{3} using \semiapproach with $\mathcal{C}^{E_3}$ labeling method.
To realize zero-shot transfer, we then utilize $\pi_L^{\pnp{1}}$ (i.e., the low-level policy of \pnp{1}) to complete \pnp{2} and \pnp{3}.
We compare the performance of this zero-shot transfer policy, with the policy trained directly on \pnp{2} and \pnp{3}.
As summarized in Table~\ref{tab:R4}, we observe that the sub-task policies from \pnp{1} (represented as $\pi_L^{\pnp{1}}$), when applied to more-complex tasks perform comparably to the policies learned specifically for those tasks (represented as $\pi_L^{\pnp{n}}$).
\textit{This exploratory analysis suggests that low-level policies learned by \approach for simpler task fit well within the hierarchical structure established by \approach~ for more complex tasks that involve similar options connotation.}

\begin{table}[t]
\centering 
\begin{tabular}{ccc} 
\toprule
\textbf{Task} &  $\pi_L^{\pnp{n}}$ & $\pi_L^{\pnp{1}}$ \\ 
\midrule
\pnp{2} & $2.9\pm 0.83$ & $2.6\pm 0.49$ \\ 
\pnp{3} & $3.5\pm 0.45$ & $3.2\pm 1.13$ \\ 
\bottomrule
\end{tabular}
\caption{\approach: Potential for Zero-Shot Transfer. Averages returns of the learned low-level policies (labeled in column) on long-horizon tasks (labeled in rows).}
\label{tab:R4}
\end{table}

\paragraph{Video Demonstrations.}
The supplementary material includes videos of both expert and learned behaviors. These videos showcase both successful instances and failure cases.
\section{Conclusion}
We introduce \approach an approach for offline IL, which estimates stationary distribution ratios to derive goal-conditioned option-aware policies.
By segmenting long-horizon demonstrations, \approach discerns a hierarchy of sub-tasks, learning distinct micro-policies for each segment and a macro-policy for sub-task transitions, while maintaining the flexibility to adapt to changing goals across different tasks.
We evaluate our approach on robotic manipulation tasks, which have been challenging for previous offline IL techniques due to their long horizon and changing goals.
Our R1 experiment showcases \approach's superior performance compared to recent offline IL baselines in these tasks.
Through experiments R2 and R3, we also evaluate the effect of expert sub-task annotations and associated hyperparameters.
Notably, our approach was able to leverage expert annotations of task segments to further enhance its learning performance.
Beyond providing a promising approach for solving long-horizon tasks, these experiments also highlight the impact of auxiliary inputs for robot learning.

\paragraph{Limitations and Future Work.}
Our work also motivates several directions of future work.
First, while \approach is able to outperform recent baselines, the performance improvement is lower as the number of objects increases.
Multiple reasons could be behind this observation, including difficulty in inferring sub-tasks over long horizons, complexity of multi-object manipulation, and need for additional training data.
Future work that investigates the underlying root cause can enhance performance of offline IL on long-horizon tasks; \approach can serve as a useful starting point for this investigation.
Second, the evaluations (though conducted on challenging long-horizon tasks) are limited to the robotics domain.
We encourage replication studies that evaluate the generality of the proposed approach on tasks derived from other domains.

Third, our experiments suggest that low-level policies learned with \approach could be used for more challenging long-horizon tasks, given that a user or a different algorithm can detail the high-level policy or sub-task sequence.
As such, a promising near-term direction is to explore the potential of \approach for offline pre-training of online IL or RL techniques that address long-horizon tasks; currently, most techniques utilize behavioral cloning for pre-training.
Finally, human-centered evaluation and subsequent development of human-guided IL techniques that utilize \approach as a subroutine are of high interest.

\paragraph{Ethical Statement.}
Ensuring safety is essential for ethical deployment of learning-based AI systems.
Our work contributes an approach to the paradigm of offline IL, which inherently facilitates safe learning by removing the requirement of (potentially unsafe) exploration.

\paragraph{Acknowledgements.}
We thank the anonymous reviewers for their detailed and constructive feedback. This research was supported in part by NSF award \#2205454, the Army Research Office through Cooperative Agreement Number W911NF-20-2-0214, and Rice University funds.

\bibliography{aaai24}

\begin{thebibliography}{45}
\providecommand{\natexlab}[1]{#1}

\bibitem[{Abbeel and Ng(2004)}]{abbeel2004apprenticeship}
Abbeel, P.; and Ng, A.~Y. 2004.
\newblock Apprenticeship learning via inverse reinforcement learning.
\newblock In \emph{Proceedings of the twenty-first international conference on Machine learning}.

\bibitem[{Andrychowicz et~al.(2017)Andrychowicz, Wolski, Ray, Schneider, Fong, Welinder, McGrew, Tobin, Pieter~Abbeel, and Zaremba}]{andrychowicz2017hindsight}
Andrychowicz, M.; Wolski, F.; Ray, A.; Schneider, J.; Fong, R.; Welinder, P.; McGrew, B.; Tobin, J.; Pieter~Abbeel, O.; and Zaremba, W. 2017.
\newblock Hindsight experience replay.
\newblock \emph{Advances in neural information processing systems}, 30.

\bibitem[{Arora and Doshi(2021)}]{arora2021survey}
Arora, S.; and Doshi, P. 2021.
\newblock A survey of inverse reinforcement learning: Challenges, methods and progress.
\newblock \emph{Artificial Intelligence}, 297: 103500.

\bibitem[{Brockman et~al.(2016)Brockman, Cheung, Pettersson, Schneider, Schulman, Tang, and Zaremba}]{brockman2016openai}
Brockman, G.; Cheung, V.; Pettersson, L.; Schneider, J.; Schulman, J.; Tang, J.; and Zaremba, W. 2016.
\newblock Openai gym.
\newblock \emph{arXiv preprint arXiv:1606.01540}.

\bibitem[{Byrne and Russon(1998)}]{byrne1998learning}
Byrne, R.~W.; and Russon, A.~E. 1998.
\newblock Learning by imitation: A hierarchical approach.
\newblock \emph{Behavioral and brain sciences}, 21(5): 667--684.

\bibitem[{Chen, Lan, and Aggarwal(2023)}]{chen2023option}
Chen, J.; Lan, T.; and Aggarwal, V. 2023.
\newblock Option-Aware Adversarial Inverse Reinforcement Learning for Robotic Control.
\newblock In \emph{2023 IEEE International Conference on Robotics and Automation (ICRA)}, 5902--5908. IEEE.

\bibitem[{Chernova and Thomaz(2014)}]{chernova2014robot}
Chernova, S.; and Thomaz, A.~L. 2014.
\newblock \emph{Robot learning from human teachers}.
\newblock Morgan \& Claypool Publishers.

\bibitem[{Daniel et~al.(2016)Daniel, Van~Hoof, Peters, and Neumann}]{daniel2016probabilistic}
Daniel, C.; Van~Hoof, H.; Peters, J.; and Neumann, G. 2016.
\newblock Probabilistic inference for determining options in reinforcement learning.
\newblock \emph{Machine Learning}, 104: 337--357.

\bibitem[{Ding et~al.(2019)Ding, Florensa, Abbeel, and Phielipp}]{ding2019goal}
Ding, Y.; Florensa, C.; Abbeel, P.; and Phielipp, M. 2019.
\newblock Goal-conditioned imitation learning.
\newblock \emph{Advances in neural information processing systems}, 32.

\bibitem[{Fang et~al.(2018)Fang, Zhou, Shi, Gong, Xu, and Zhang}]{fang2018dher}
Fang, M.; Zhou, C.; Shi, B.; Gong, B.; Xu, J.; and Zhang, T. 2018.
\newblock DHER: Hindsight experience replay for dynamic goals.
\newblock In \emph{International Conference on Learning Representations}.

\bibitem[{Gao, Jiang, and Chen(2023)}]{gao2023transferring}
Gao, C.; Jiang, Y.; and Chen, F. 2023.
\newblock Transferring hierarchical structures with dual meta imitation learning.
\newblock In \emph{Conference on Robot Learning}, 762--773. PMLR.

\bibitem[{Gupta et~al.(2019)Gupta, Kumar, Lynch, Levine, and Hausman}]{gupta2019relay}
Gupta, A.; Kumar, V.; Lynch, C.; Levine, S.; and Hausman, K. 2019.
\newblock Relay policy learning: Solving long-horizon tasks via imitation and reinforcement learning.
\newblock \emph{arXiv preprint arXiv:1910.11956}.

\bibitem[{Habibian, Jonnavittula, and Losey(2022)}]{habibian2022here}
Habibian, S.; Jonnavittula, A.; and Losey, D.~P. 2022.
\newblock Here’s what I’ve learned: Asking questions that reveal reward learning.
\newblock \emph{ACM Transactions on Human-Robot Interaction (THRI)}, 11(4): 1--28.

\bibitem[{Ho and Ermon(2016)}]{ho2016generative}
Ho, J.; and Ermon, S. 2016.
\newblock Generative adversarial imitation learning.
\newblock \emph{Advances in neural information processing systems}, 29.

\bibitem[{Jamgochian et~al.(2023)Jamgochian, Buehrle, Fischer, and Kochenderfer}]{jamgochian2023shail}
Jamgochian, A.; Buehrle, E.; Fischer, J.; and Kochenderfer, M.~J. 2023.
\newblock SHAIL: Safety-Aware Hierarchical Adversarial Imitation Learning for Autonomous Driving in Urban Environments.
\newblock In \emph{2023 IEEE International Conference on Robotics and Automation (ICRA)}, 1530--1536. IEEE.

\bibitem[{Jing et~al.(2021)Jing, Huang, Sun, Ma, Kong, Gan, and Li}]{jing2021adversarial}
Jing, M.; Huang, W.; Sun, F.; Ma, X.; Kong, T.; Gan, C.; and Li, L. 2021.
\newblock Adversarial option-aware hierarchical imitation learning.
\newblock In \emph{International Conference on Machine Learning}, 5097--5106. PMLR.

\bibitem[{Kim et~al.(2022)Kim, Lee, Jang, Yang, and Kim}]{kim2022lobsdice}
Kim, G.-H.; Lee, J.; Jang, Y.; Yang, H.; and Kim, K.-E. 2022.
\newblock LobsDICE: Offline Learning from Observation via Stationary Distribution Correction Estimation.
\newblock \emph{Advances in Neural Information Processing Systems}, 35: 8252--8264.

\bibitem[{Kim et~al.(2021)Kim, Seo, Lee, Jeon, Hwang, Yang, and Kim}]{kim2021demodice}
Kim, G.-H.; Seo, S.; Lee, J.; Jeon, W.; Hwang, H.; Yang, H.; and Kim, K.-E. 2021.
\newblock Demodice: Offline imitation learning with supplementary imperfect demonstrations.
\newblock In \emph{International Conference on Learning Representations}.

\bibitem[{Kostrikov, Nachum, and Tompson(2019)}]{kostrikov2019imitation}
Kostrikov, I.; Nachum, O.; and Tompson, J. 2019.
\newblock Imitation learning via off-policy distribution matching.
\newblock \emph{arXiv preprint arXiv:1912.05032}.

\bibitem[{Le et~al.(2018)Le, Jiang, Agarwal, Dud{\'\i}k, Yue, and Daum{\'e}~III}]{le2018hierarchical}
Le, H.; Jiang, N.; Agarwal, A.; Dud{\'\i}k, M.; Yue, Y.; and Daum{\'e}~III, H. 2018.
\newblock Hierarchical imitation and reinforcement learning.
\newblock In \emph{International conference on machine learning}, 2917--2926. PMLR.

\bibitem[{Lee et~al.(2021)Lee, Jeon, Lee, Pineau, and Kim}]{lee2021optidice}
Lee, J.; Jeon, W.; Lee, B.; Pineau, J.; and Kim, K.-E. 2021.
\newblock Optidice: Offline policy optimization via stationary distribution correction estimation.
\newblock In \emph{International Conference on Machine Learning}, 6120--6130. PMLR.

\bibitem[{Li, Song, and Ermon(2017)}]{li2017infogail}
Li, Y.; Song, J.; and Ermon, S. 2017.
\newblock Infogail: Interpretable imitation learning from visual demonstrations.
\newblock \emph{Advances in neural information processing systems}, 30.

\bibitem[{Ma et~al.(2022{\natexlab{a}})Ma, Yan, Jayaraman, and Bastani}]{ma2022offline}
Ma, J.~Y.; Yan, J.; Jayaraman, D.; and Bastani, O. 2022{\natexlab{a}}.
\newblock Offline goal-conditioned reinforcement learning via $ f $-advantage regression.
\newblock \emph{Advances in Neural Information Processing Systems}, 35: 310--323.

\bibitem[{Ma et~al.(2022{\natexlab{b}})Ma, Shen, Jayaraman, and Bastani}]{ma2022versatile}
Ma, Y.; Shen, A.; Jayaraman, D.; and Bastani, O. 2022{\natexlab{b}}.
\newblock Versatile offline imitation from observations and examples via regularized state-occupancy matching.
\newblock In \emph{International Conference on Machine Learning}, 14639--14663. PMLR.

\bibitem[{Nachum et~al.(2019)Nachum, Dai, Kostrikov, Chow, Li, and Schuurmans}]{nachum2019algaedice}
Nachum, O.; Dai, B.; Kostrikov, I.; Chow, Y.; Li, L.; and Schuurmans, D. 2019.
\newblock Algaedice: Policy gradient from arbitrary experience.
\newblock \emph{arXiv preprint arXiv:1912.02074}.

\bibitem[{Nasiriany et~al.(2023)Nasiriany, Gao, Mandlekar, and Zhu}]{nasiriany2023learning}
Nasiriany, S.; Gao, T.; Mandlekar, A.; and Zhu, Y. 2023.
\newblock Learning and Retrieval from Prior Data for Skill-based Imitation Learning.
\newblock In \emph{Conference on Robot Learning}, 2181--2204. PMLR.

\bibitem[{Orlov-Savko et~al.(2022)Orlov-Savko, Jain, Gremillion, Neubauer, Canady, and Unhelkar}]{orlov2022factorial}
Orlov-Savko, L.; Jain, A.; Gremillion, G.~M.; Neubauer, C.~E.; Canady, J.~D.; and Unhelkar, V. 2022.
\newblock Factorial Agent Markov Model: Modeling Other Agents' Behavior in presence of Dynamic Latent Decision Factors.
\newblock In \emph{Proceedings of the 21st International Conference on Autonomous Agents and Multiagent Systems}, 982--990.

\bibitem[{Osa et~al.(2018)Osa, Pajarinen, Neumann, Bagnell, Abbeel, Peters et~al.}]{osa2018algorithmic}
Osa, T.; Pajarinen, J.; Neumann, G.; Bagnell, J.~A.; Abbeel, P.; Peters, J.; et~al. 2018.
\newblock An algorithmic perspective on imitation learning.
\newblock \emph{Foundations and Trends{\textregistered} in Robotics}, 7(1-2): 1--179.

\bibitem[{Plappert et~al.(2018)Plappert, Andrychowicz, Ray, McGrew, Baker, Powell, Schneider, Tobin, Chociej, Welinder et~al.}]{plappert2018multi}
Plappert, M.; Andrychowicz, M.; Ray, A.; McGrew, B.; Baker, B.; Powell, G.; Schneider, J.; Tobin, J.; Chociej, M.; Welinder, P.; et~al. 2018.
\newblock Multi-goal reinforcement learning: Challenging robotics environments and request for research.
\newblock \emph{arXiv preprint arXiv:1802.09464}.

\bibitem[{Pomerleau(1991)}]{pomerleau1991efficient}
Pomerleau, D.~A. 1991.
\newblock Efficient training of artificial neural networks for autonomous navigation.
\newblock \emph{Neural computation}, 3(1): 88--97.

\bibitem[{Puterman(2014)}]{puterman2014markov}
Puterman, M.~L. 2014.
\newblock \emph{Markov decision processes: discrete stochastic dynamic programming}.
\newblock John Wiley \& Sons.

\bibitem[{Quintero-Pena et~al.(2022)Quintero-Pena, Chamzas, Sun, Unhelkar, and Kavraki}]{quintero2022human}
Quintero-Pena, C.; Chamzas, C.; Sun, Z.; Unhelkar, V.; and Kavraki, L.~E. 2022.
\newblock Human-guided motion planning in partially observable environments.
\newblock In \emph{2022 International Conference on Robotics and Automation (ICRA)}, 7226--7232. IEEE.

\bibitem[{Ranchod, Rosman, and Konidaris(2015)}]{ranchod2015nonparametric}
Ranchod, P.; Rosman, B.; and Konidaris, G. 2015.
\newblock Nonparametric bayesian reward segmentation for skill discovery using inverse reinforcement learning.
\newblock In \emph{2015 IEEE/RSJ International Conference on Intelligent Robots and Systems (IROS)}, 471--477. IEEE.

\bibitem[{Ravichandar et~al.(2020)Ravichandar, Polydoros, Chernova, and Billard}]{ravichandar2020recent}
Ravichandar, H.; Polydoros, A.~S.; Chernova, S.; and Billard, A. 2020.
\newblock Recent advances in robot learning from demonstration.
\newblock \emph{Annual review of control, robotics, and autonomous systems}, 3: 297--330.

\bibitem[{Ross, Gordon, and Bagnell(2011)}]{ross2011reduction}
Ross, S.; Gordon, G.; and Bagnell, D. 2011.
\newblock A reduction of imitation learning and structured prediction to no-regret online learning.
\newblock In \emph{Proceedings of the fourteenth international conference on artificial intelligence and statistics}, 627--635. JMLR Workshop and Conference Proceedings.

\bibitem[{Schaul et~al.(2015)Schaul, Horgan, Gregor, and Silver}]{schaul2015universal}
Schaul, T.; Horgan, D.; Gregor, K.; and Silver, D. 2015.
\newblock Universal value function approximators.
\newblock In \emph{International conference on machine learning}, 1312--1320. PMLR.

\bibitem[{Seo and Unhelkar(2022)}]{seo2022semi}
Seo, S.; and Unhelkar, V. 2022.
\newblock Semi-Supervised Imitation Learning of Team Policies from Suboptimal Demonstrations.
\newblock In \emph{International Joint Conference on Artificial Intelligence (IJCAI)}.

\bibitem[{Sharma et~al.(2018)Sharma, Sharma, Rhinehart, and Kitani}]{sharma2018directed}
Sharma, A.; Sharma, M.; Rhinehart, N.; and Kitani, K.~M. 2018.
\newblock Directed-info gail: Learning hierarchical policies from unsegmented demonstrations using directed information.
\newblock \emph{arXiv preprint arXiv:1810.01266}.

\bibitem[{Sutton, Precup, and Singh(1999)}]{sutton1999between}
Sutton, R.~S.; Precup, D.; and Singh, S. 1999.
\newblock Between MDPs and semi-MDPs: A framework for temporal abstraction in reinforcement learning.
\newblock \emph{Artificial intelligence}, 112(1-2): 181--211.

\bibitem[{Todorov, Erez, and Tassa(2012)}]{todorov2012mujoco}
Todorov, E.; Erez, T.; and Tassa, Y. 2012.
\newblock Mujoco: A physics engine for model-based control.
\newblock In \emph{2012 IEEE/RSJ international conference on intelligent robots and systems}, 5026--5033. IEEE.

\bibitem[{Unhelkar, Li, and Shah(2020)}]{unhelkar2020semi}
Unhelkar, V.~V.; Li, S.; and Shah, J.~A. 2020.
\newblock Semi-supervised learning of decision-making models for human-robot collaboration.
\newblock In \emph{Conference on Robot Learning}, 192--203. PMLR.

\bibitem[{Unhelkar and Shah(2019)}]{unhelkar2019learning}
Unhelkar, V.~V.; and Shah, J.~A. 2019.
\newblock Learning models of sequential decision-making with partial specification of agent behavior.
\newblock In \emph{Proceedings of the AAAI conference on artificial intelligence}, volume~33, 2522--2530.

\bibitem[{Wang et~al.(2021)Wang, Xu, Du, and Lee}]{wang2021learning}
Wang, Y.; Xu, C.; Du, B.; and Lee, H. 2021.
\newblock Learning to weight imperfect demonstrations.
\newblock In \emph{International Conference on Machine Learning}, 10961--10970. PMLR.

\bibitem[{Wu et~al.(2019)Wu, Charoenphakdee, Bao, Tangkaratt, and Sugiyama}]{wu2019imitation}
Wu, Y.-H.; Charoenphakdee, N.; Bao, H.; Tangkaratt, V.; and Sugiyama, M. 2019.
\newblock Imitation learning from imperfect demonstration.
\newblock In \emph{International Conference on Machine Learning}, 6818--6827. PMLR.

\bibitem[{Wu et~al.(2021)Wu, Lian, Unhelkar, Tomizuka, and Schaal}]{wu2021learning}
Wu, Z.; Lian, W.; Unhelkar, V.; Tomizuka, M.; and Schaal, S. 2021.
\newblock Learning dense rewards for contact-rich manipulation tasks.
\newblock In \emph{2021 IEEE International Conference on Robotics and Automation (ICRA)}, 6214--6221. IEEE.

\end{thebibliography}

\section{Appendix}

\begin{table*}
    \centering
    \small\scshape 
    \begin{tabular}{|c|c|c|c|c|c|}
        \hline
        Task & Illustration & State Dimension & Goal Dimension & Action Dimension & Horizon, $T$ \\
        \hline
        \pnp{1} & \includegraphics[width=2.2cm]{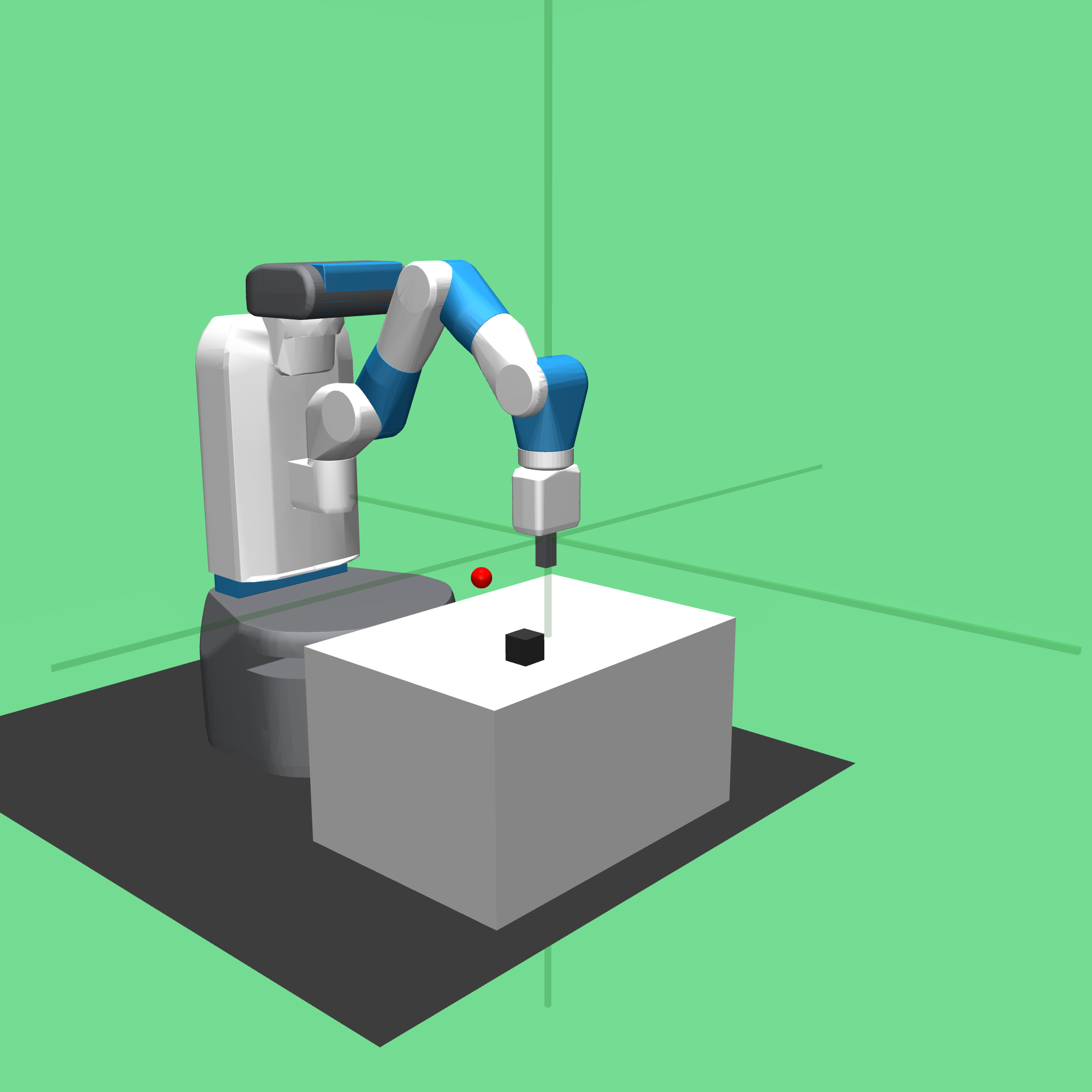} & 10 & 3 & 4 & 100 \\
        \hline
        \pnp{2} & \includegraphics[width=2.2cm]{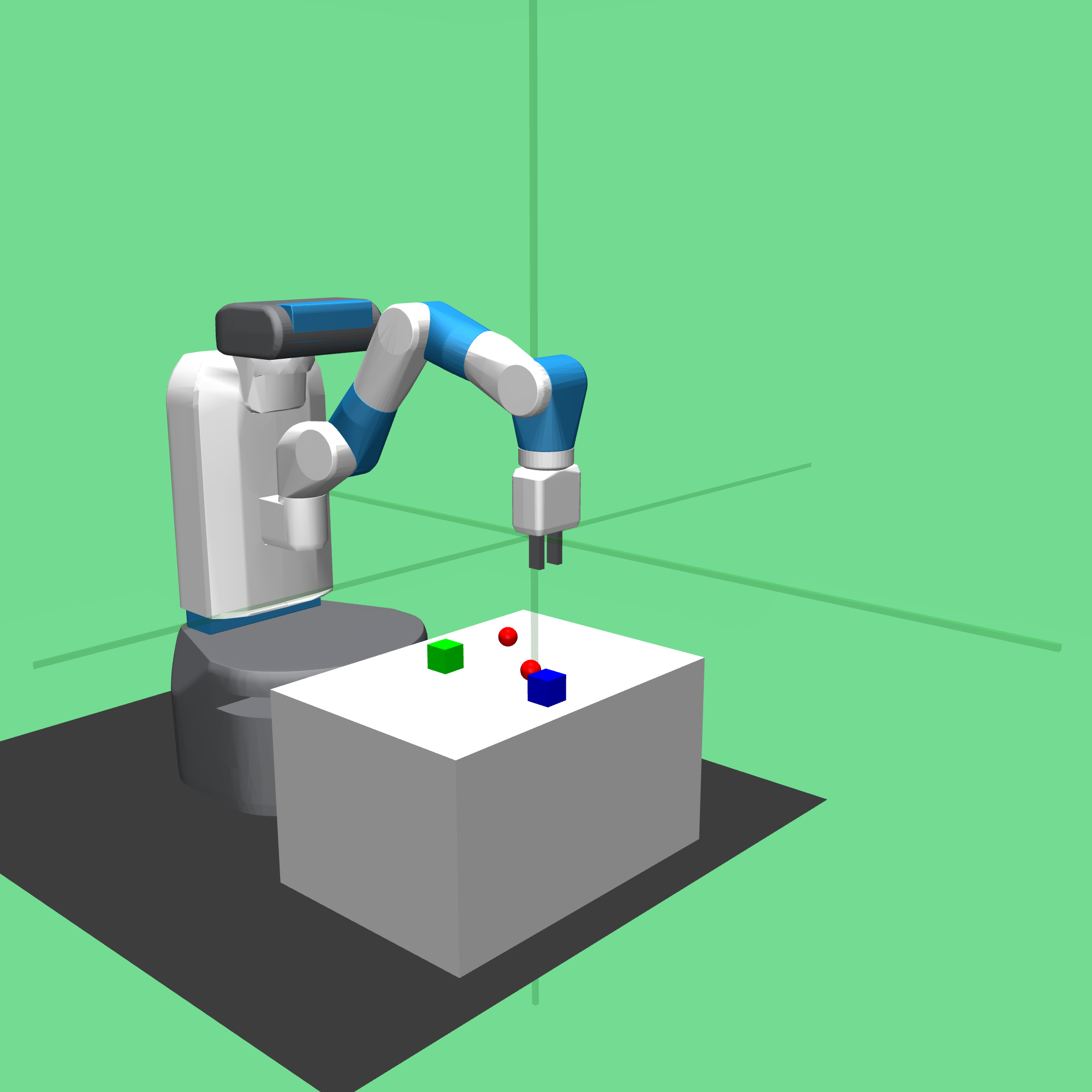} & 16 & 6 & 4 & 150 \\
        \hline
        \pnp{3} & \includegraphics[width=2.2cm]{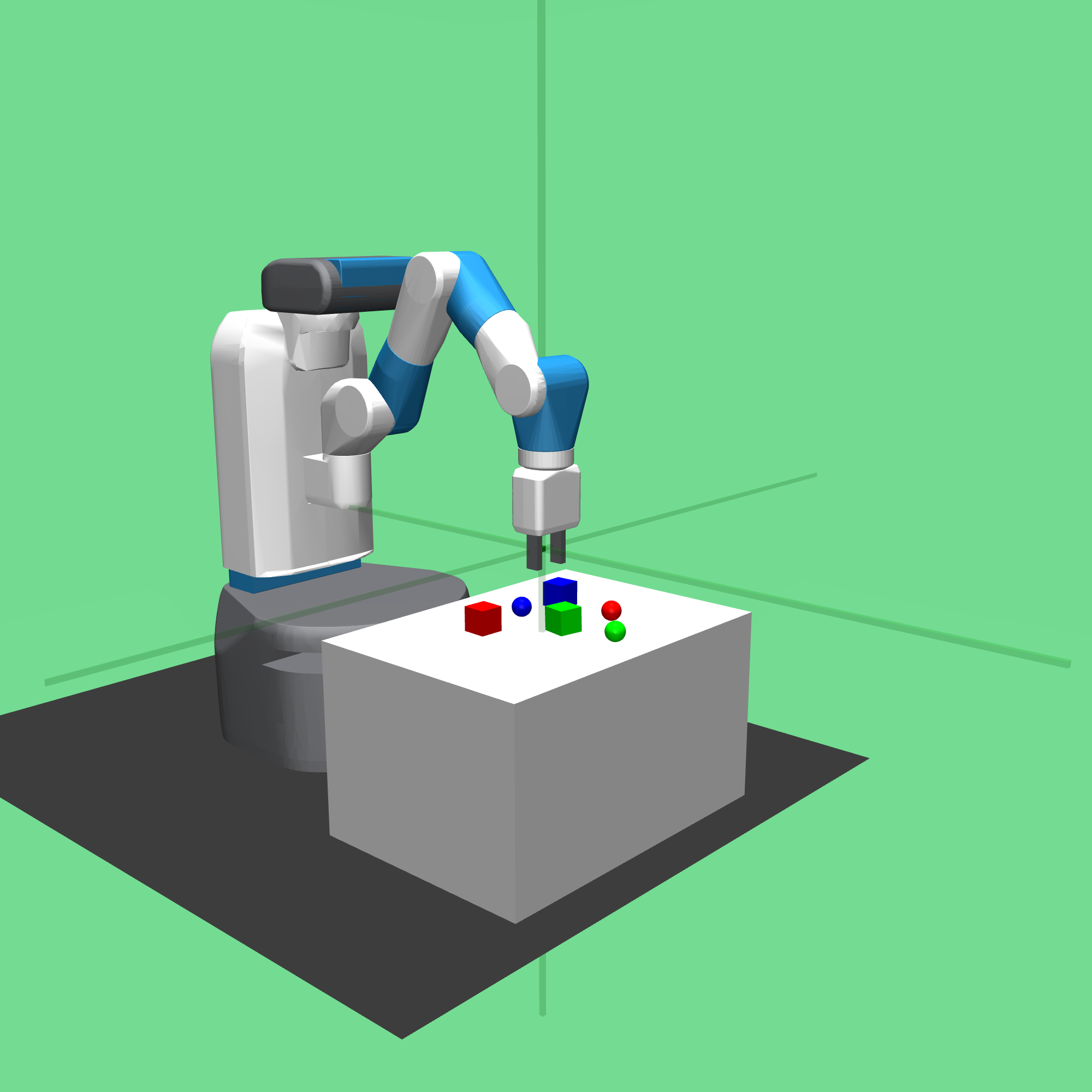} & 22 & 9 & 4 & 250 \\
        \hline
        \stack{3} & \includegraphics[width=2.2cm]{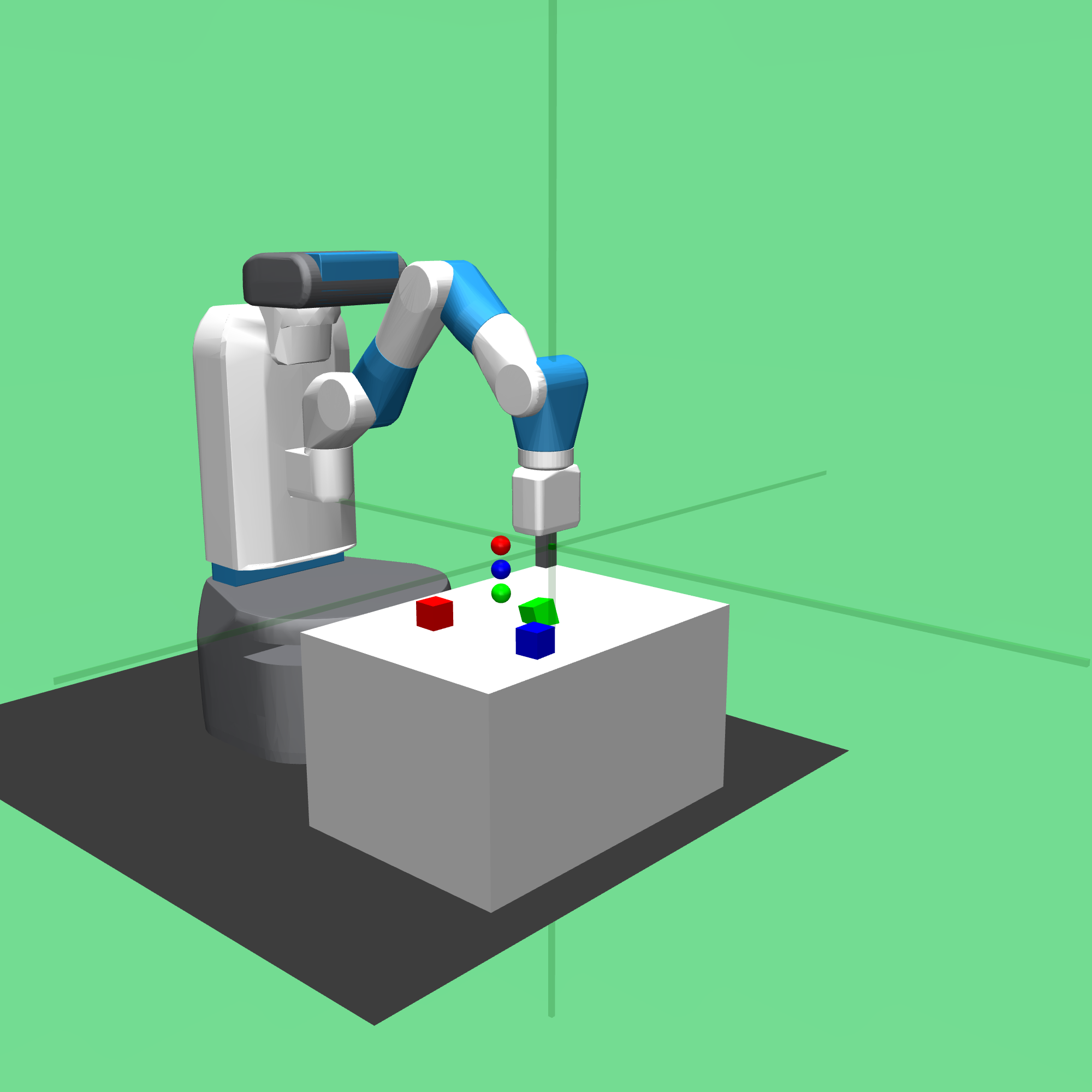} & 22 & 9 & 4 & 200 \\
        \hline
    \end{tabular}
    \label{tab:task_details}
    \caption{Task configuration of \pnp{n} for $n\in \{1,2,3\}$ and \texttt{Stack}-3 tasks.}
    \vspace{10em}
\end{table*}

\begin{table*}
    \centering
    \small\scshape 
    \begin{tabular}{lll}
        \toprule
        \textbf{Category} & \textbf{Hyperparameter} & \textbf{Value} \\
        \midrule
        \multirow{7}{*}{ADAM Optimizer} & $\beta_1$ & 0.9 \\
        & $\beta_2$ & 0.999 \\
        & $\epsilon$ & 1e-7 \\
        & Learning Rate (Critic) &  3e-4\\
        & Learning Rate (Discriminator) &  3e-4\\
        & Learning Rate (Low-Level Policy) &  3e-3\\
        & Learning Rate (High-Level Policy)&  3e-3\\
        \midrule
        \multirow{6}{*}{Architecture} & Critic $(\nu)$ &  [256, 256, 128]\\
            & Discriminator $(\Psi)$ & [256, 256, 128] \\
            & Low-Level Policy $(\pi_L)$ & [256, 256, 128] \\
            & High-Level Policy $(\pi_H)$ &  [256, 256, 128]\\
            & Activation function (all) & ReLU \\
        \midrule
        \multirow{5}{*}{Training} & Batch Size & 256$\times n_{objects}$ \\
        & Iterations & 10000 \\
        & Gradient Penalty Coeff. $(\Psi)$ & 10 \\
        & Gradient Penalty Coeff. $(\nu)$ & 1e-4 \\
        & Discount, $\gamma$ & 0.99 \\
        \bottomrule
    \end{tabular}
    \caption{Hyper-parameter configuration used for our experiments.}
    \label{tab:hyperparam}
\end{table*}

\begin{figure*}
\centering
\begin{subfigure}{.249\textwidth}
  \centering
  \includegraphics[width=\linewidth]{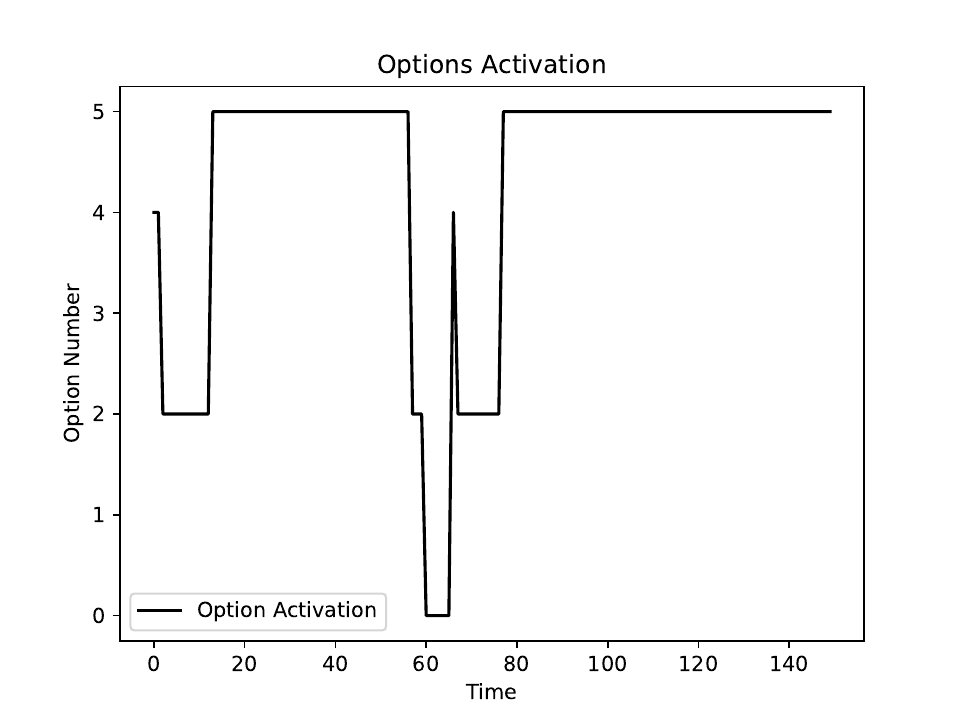}
  \caption{\pnp{2}; \approach}
  \label{subfig:PnPx2_GODICE_option}
\end{subfigure}%
\hfill
\begin{subfigure}{.249\textwidth}
  \centering
  \includegraphics[width=\linewidth]{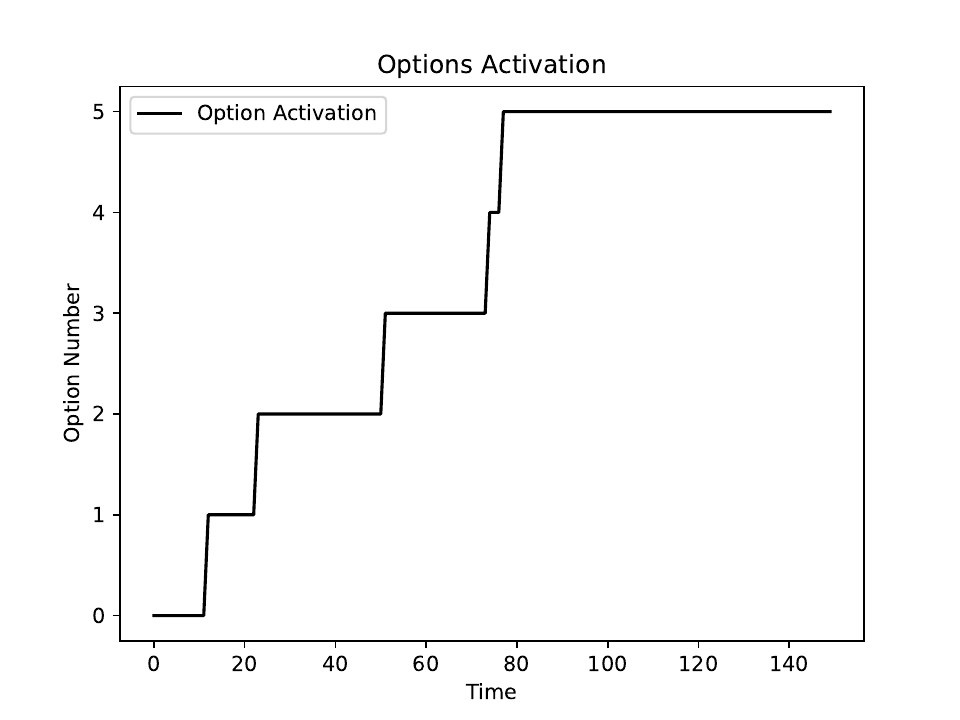}
  \caption{\pnp{2}; \semiapproach}
  \label{subfig:PnPx2_GODICE_Semi_option}
\end{subfigure}%
\hfill
\begin{subfigure}{.249\textwidth}
  \centering
  \includegraphics[width=\linewidth]{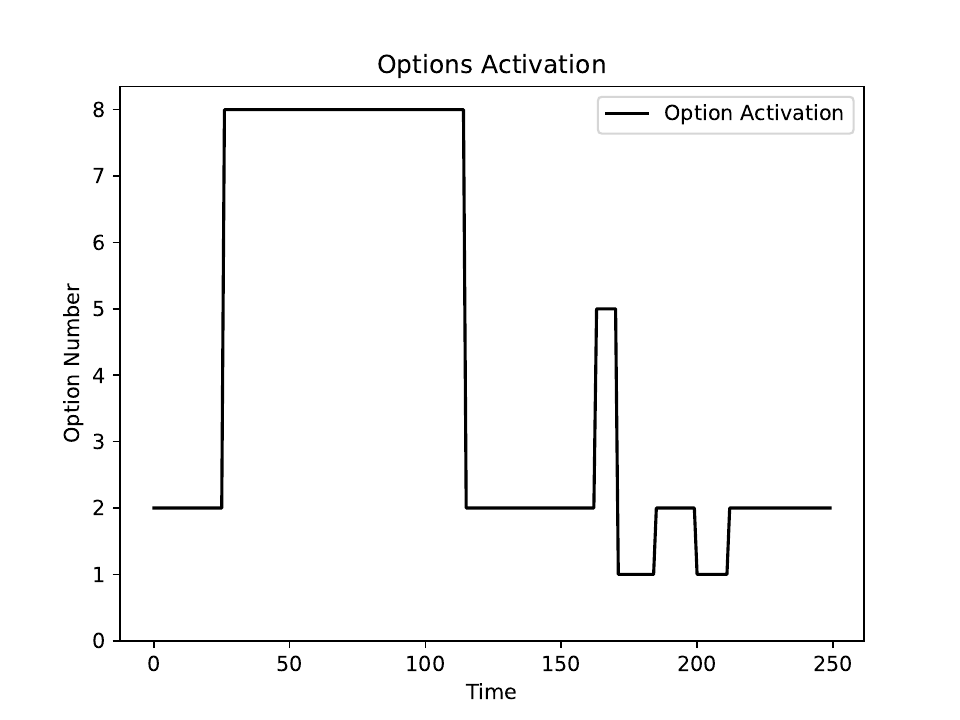}
  \caption{\pnp{3}; \approach}
  \label{subfig:PnPx3_GODICE_option}
\end{subfigure}%
\hfill
\begin{subfigure}{.249\textwidth}
  \centering
  \includegraphics[width=\linewidth]{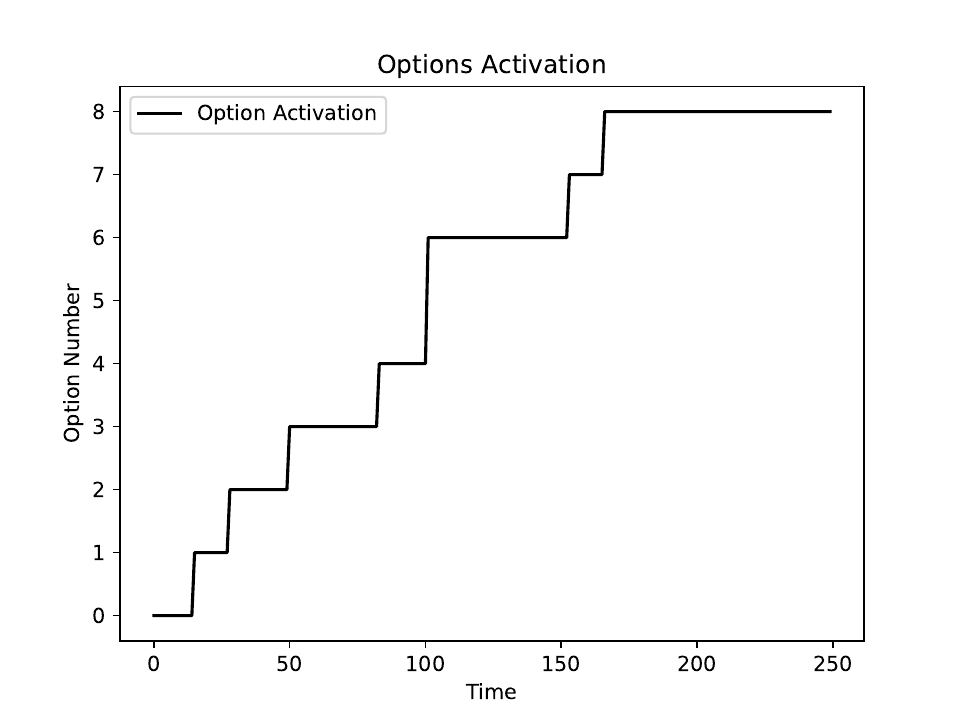}
  \caption{\pnp{3}; \semiapproach}
  \label{subfig:PnPx3_GODICE_Semi_option}
\end{subfigure}%
\caption{Figure illustrating the activation of options over time by \approach and \semiapproach for a sample episode in both \pnp{2} and \pnp{3} tasks. $y$-axis denotes each discrete option and $x$-axis denotes the task horizon. Note that $c\in\{1,3\}$ and $c\in\{0,3,4,6,7\}$ are not activated by \approach in \pnp{2}(a) and \pnp{3}(c) respectively.}
\label{fig:OptionActivation}
\end{figure*}

The following appendices provide additional details regarding the algorithm, experiments, and results.

\subsection{Derivation of Direct Convex Optimization (Eq.~\ref{eq:value_func_objective_unstable})}

Let us first restate the dual formulation Eq. \ref{eq:lagrangian function} we obtained
\begin{align}
     &\max_{d^{{\pi}}\geq0}\min_{\nu}f(\nu, d) \; {-} \; D_{KL}(d^{{\pi}}||d^{\pi_E}) \; {-} \; \alpha D_{KL}(d^{{\pi}}||d^{\pi_O}) \nonumber \\
     &\text{where} \; f(\nu, d^{{\pi}}) = \sum_{c',s,g}\nu(c',s,g)\bigg((1-\gamma){\mu}(c', s; g) +  \nonumber \\
     &\phantom{f(\nu, d^{{\pi}}) = \sum_{c',s,g}} \gamma (T_*d^{{\pi}})(c',s, g) - \sum_{c, a} d^{\pi}(c', s, c, a; g)\bigg), \nonumber
\end{align}%
Now, let us simplify $f$ by rearranging its terms
\begin{align}
     f(\nu, d^{{\pi}}) =&  (1-\gamma)\sum_{c',s,g}{\mu}(c', s; g)\nu(c',s,g) + \nonumber\\
     & \sum_{c',s,c,a,g}d^{\pi}(.)\Big(\gamma\sum_{s'}\mathbf{T}(s'|s, a)\nu(c, s', g)\nonumber\\
    &\phantom{+ \sum_{c',s,c,a,g}d^{\pi}(.)\Big(}  - \nu(c', s, g)\Big)  \nonumber\\
    =& (1-\gamma)\mathbb{E}_{\mu}[\nu(.)] + \mathbb{E}_{d^{\pi}}[\gamma (T\nu)(s, c, a, g) {-} \nu(.)]\nonumber
\end{align}%
Plugging this back in original formulation while using the definition of KL-divergence, we obtain the following
\begin{align}
     &\max_{d^{{\pi}}\geq0}\min_{\nu} (1-\gamma)\mathbb{E}_{\mu}[\nu(.)] +  \mathbb{E}_{d^{\pi}}[\log\frac{d^{\pi_E}}{d^{\pi_O}} + \gamma (T\nu){-} \nu(.) \nonumber\\
     &\phantom{\max_{d^{{\pi}}\geq0}\min_{\nu} (1-\gamma)\mathbb{E}_{\mu}[\nu(.)] +  \mathbb{E}_{d^{\pi}}} {-} (1+\alpha)\log\frac{d^{\pi}}{d^{\pi_O}}] \nonumber
\end{align}%
With the introduced stationary distribution ratio $w$ (Eq. \ref{eq:imp_weight}) and advantage function $A_\nu$ (Eq. \ref{eq:advantage_fn}-\ref{eq:log_distribution_ratio}), we finally obtain the maxmin optimization problem of Eq. \ref{eq:new lagrangian function}
\begin{align}
   \max_{w\geq 0}&\min_{\nu} \; (1-\gamma)\mathbb{E}_{{\mu}}[\nu(.)] + \mathbb{E}_{d^{\pi_O}}\Big[w(c',s,c,a, g) \nonumber \\
   &\big(A_{\nu}(c',s,c,a, g) - (1+\alpha)\log w(c',s,c,a, g)\big)\Big]\nonumber
\end{align}
By observing that the original objective Eq. \ref{eq:il_dist_match_reg} is a convex optimization problem \cite{lee2021optidice} and using the assumption that there exists $d_{\pi}(c's,c,a;g)$ that satisfies the Bellman constraints make the problem strictly feasible and, thus, strong duality holds. Therefore, we can interchange the order of optimization to obtain the following
\begin{align}
   \min_{\nu}&\max_{w\geq 0} \; (1-\gamma)\mathbb{E}_{{\mu}}[\nu(.)] + \mathbb{E}_{d^{\pi_O}}\Big[w(c',s,c,a, g) \nonumber \\
   &\big(A_{\nu}(c',s,c,a, g) - (1+\alpha)\log w(c',s,c,a, g)\big)\Big]\nonumber
\end{align}
Now, by solving the inner maximum optimization 
$$\arg\max_w\mathbb{E}_{d^{\pi_O}}\Big[w(.)\big(A_{\nu}- (1+\alpha)\log w(.)\big)\Big]$$
we can obtain $w^*=\exp\big(A_{\nu}/(1{+}\alpha){-}1\big)$ (Eq. \ref{eq:optimal_imp_weight}).
The resulting minmax formulation can be reduced to a single unconstrained minimum optimization problem over $\nu$ (Eq. \ref{eq:value_func_objective_unstable})
\begin{equation}
    \min_{\nu} \; (1-\gamma)\mathbb{E}_{{\mu}}[\nu(.)] {+} (1+\alpha)\mathbb{E}_{d^{\pi_O}}[w^*(\cdot)]\nonumber
\end{equation}

\begin{table*}
    \centering 
     \scshape
        \begin{tabular}{ccccccc} 
            \toprule
            \textbf{Model} & $c=0$ & $c=1$ & $c=2$ & $c=3$ & $c=4$ & $c=5$\\
            \midrule
            \approach & $0.31\pm0.34$ & $0.08\pm0.12$ & $0.12\pm0.13$ & $0.07\pm0.12$ & $0.12\pm0.19$ & $0.71\pm0.27$\\ 
            \semiapproach & $0.11\pm0.10$ & $0.27\pm0.26$ & $0.16\pm0.16$ &  $0.30\pm0.22$ & $0.30\pm0.22$ & $0.19\pm0.20$\\ 
            Expert, $E_3$ & $0.10\pm0.10$ & $0.08\pm0.01$ & $0.17\pm0.07$ & $0.20\pm0.07$ & $0.08\pm0.04$ & $0.41\pm0.04$\\ 
            \bottomrule
        \end{tabular}
    \caption{Average fraction of time spent in each option for the \pnp{2} task (with a horizon of $T=150$) over 100 episodes. Both \approach and \semiapproach use $K=6$ options, where the annotations for the latter are provided by $E_3$.}
    \label{tab:OptionActivationPnPx2}
\end{table*}

\begin{table*}
    \centering 
     \scshape 
    \resizebox{\textwidth}{!}{
        \begin{tabular}{cccccccccc} 
            \toprule
            \textbf{Model} & $c=0$ & $c=1$ & $c=2$ & $c=3$ & $c=4$ & $c=5$ & $c=6$ & $c=7$ & $c=8$ \\
            \midrule
            \approach & $-$ & $0.38\pm0.35$ & $0.48\pm0.41$ & $-$ & $-$ & $0.13\pm0.25$ & $0.03\pm0.03$ & $-$ & $0.71\pm0.33$\\ 
            \semiapproach & $0.06\pm0.00$ & $0.18\pm0.30$ & $0.27\pm0.30$ & $0.38\pm0.32$ & $0.25\pm0.26$ & $0.19\pm0.25$ & $0.46\pm0.24$ & $0.21\pm0.27$ & $0.11\pm0.12$\\ 
            Expert, $E_3$  & $0.09\pm0.11$ & $0.05\pm0.03$ & $0.21\pm0.27$ & $0.12\pm0.06$ & $0.07\pm0.10$ & $0.13\pm0.12$ & $0.10\pm0.01$ & $0.05\pm0.05$ & $0.34\pm0.12$\\
            \bottomrule
        \end{tabular}
    }
    \caption{Average fraction of time spent in each option for the \pnp{3} task (with a horizon of $T=250$) over 100 episodes. $(-)$ represents the corresponding option was never activated by the model. Both \approach and \semiapproach use $K=9$ options, where the annotations for the latter are provided by $E_3$.}
    \label{tab:OptionActivationPnPx3}
\end{table*}

Similar to \cite{kim2021demodice}, we identified potential instability during optimization over $\nu$ due to the unbounded nature of the exponential term in $w^*$. This can lead to exploding gradients. To combat this, \cite{kim2021demodice} introduced a surrogate convex objective that's both numerically stable and shares the same optimal value with the original, primarily because its gradient is a bounded softmax function. This alternative objective can be derived using Fenchel Duality \cite{ma2022versatile}, where the Fenchel conjugate of the KL-divergence results in the observed log-sum-exp function.
\begin{equation}
    \min_{\nu} \; (1-\gamma)\mathbb{E}_{{\mu}}[\nu(\cdot)] {+} (1+\alpha)\log\mathbb{E}_{d^{\pi_O}}[\exp(\frac{A_{\nu}}{1+\alpha})]
    \nonumber
\end{equation}

\section{Experiments: Implementation Details}

\subsection{Experimental Tasks}
To implement the experiments, we utilized the original one-object Fetch Pick-n-Place from Gymnasium-Robotics\footnote{\url{https://robotics.farama.org/envs/fetch/pick_and_place/}} and the Mujoco simulator~\cite{todorov2012mujoco, brockman2016openai}.
We expanded this task to include multi-object scenarios.
We consider three variants of this benchmark task -- \pnp{1}, \pnp{2}, \pnp{3} -- which include $1, 2,$ and $3$ objects, respectively.

The task complexity increases with the number of objects, requiring increasing levels of long-horizon reasoning.
Further adding to the complexity, desired goals (place locations of objects) changes across demonstrations.
These tasks are chosen for evaluation due to the fact that even the simplest variant \pnp{1} is challenging for offline IL algorithms~\cite{ding2019goal}.
Moreover, \texttt{PnP} family of tasks naturally encapsulate other primitive tasks commonly used in IL benchmarking, such as reach and grasp; thus, success in \texttt{PnP} requires the learner to also succeed in these primitive tasks.
Finally, by varying the number of objects, these tasks allows us to isolate the challenge of long horizon.

Detailed configurations, including dimension of the state space and task horizon, for the experimental tasks are summarized in Table 3.
Video demonstrations of these tasks can be found in accompanying multimedia files.

\subsection{Baselines}
We briefly compare the objectives of baselines considered for benchmarking.
\begin{align*}
    &\max_{\pi} J_{BC} = \beta \; \mathbb{E}_{(s,a,g)\sim\mathcal{D}_{O}}[\log\pi(a|s,g)] \nonumber\\
    &\phantom{\min_{\pi} J_{BC} =} {+} (1-\beta) \; \mathbb{E}_{(s,a,g)\sim\mathcal{D}_{E}}[\log\pi(a|s,g)]\\
    &\max_{d^{\pi}} J_{GoFAR} = \mathbb{E}_{g\sim p(g), s\sim d^{\pi}(.|g)}[\log\frac{d^{\pi_E}(s|g)}{d^{\pi_O}(s|g)}]\nonumber\\
    &\phantom{\max_{d^{\pi}} J_{GoFAR} = } {-} D_{\chi^2}(d^{\pi}(s,a;g)||d^{\pi_O}(s,a;g))\\
    &\max_{d^{\pi}} J_{g-DemoDICE} = {-}D_{KL}(d^{\pi}(s,a;g)||d^{\pi_E}(s,a;g))\nonumber\\
    &\phantom{\max_{d^{\pi}} J_{g-DemoDICE} = } {-} \alpha D_{KL}(d^{\pi}(s,a;g)||d^{\pi_O}(s,a;g))
\end{align*}
where for each baseline $\pi$ represents a single actor.
To ensure an equitable comparison, every baseline employs the same learning rate and actor architecture, mirroring those utilized for each option in \approach.
Details are provided in Table \ref{tab:hyperparam}.
Based on our experimental evaluations for Behavior Cloning, we determined that setting \(\beta=0.0\), which corresponds to supervised learning solely on expert demonstrations, yielded the best results.
An important observation is that \textsc{g-DemoDICE} effectively functions as a single-option version of \approach, setting $K=1$.
For this baseline, the relationship between the policies can be defined as: $\pi_H(c|s,c',g)=1(c=c')$ and $\pi_L(a|s,c,g)=\pi_L(a|s,g)$.

\section{Experiments: Additional Results}

\subsection{Explaining R2 with Option Activation}
In this section, we provide further evidence supporting our observations from experiment R2.
In Fig.~\ref{fig:R2}, it is evident that merely augmenting the number of options does not enhance \approach's performance.
For a deeper understanding, we analyzed the average fraction of time the model uses each option when a higher option count is considered, i.e., $K=6$ for \pnp{2} and $K=9$ for \pnp{3}.
This is further contrasted with \semiapproach, which benefits from expert-provided task segment annotations.

The results of these experiments are summarized in Tables~\ref{tab:OptionActivationPnPx2}-\ref{tab:OptionActivationPnPx3}. 
For \pnp{2}, \approach appears to unevenly distribute its attention, favoring options $c\in\{0, 2, 4, 5\}$.
Interestingly, for \pnp{3}, it completely omits options $c\in\{0,3,4,7\}$.
This indicates that \approach might be utilizing fewer effective options than provided to accomplish the task, shedding light on the patterns seen in Fig.~\ref{fig:R2}.
When aided by task segment annotations, \semiapproach manages to activate all options, more closely mirroring the distribution set by the Expert $E_3$.
This distinction is further illustrated in the option activation graph, as seen in Fig.~\ref{fig:OptionActivation}.

\begin{figure}[t]
\begin{subfigure}{0.56\linewidth}
  \includegraphics[width=\linewidth]{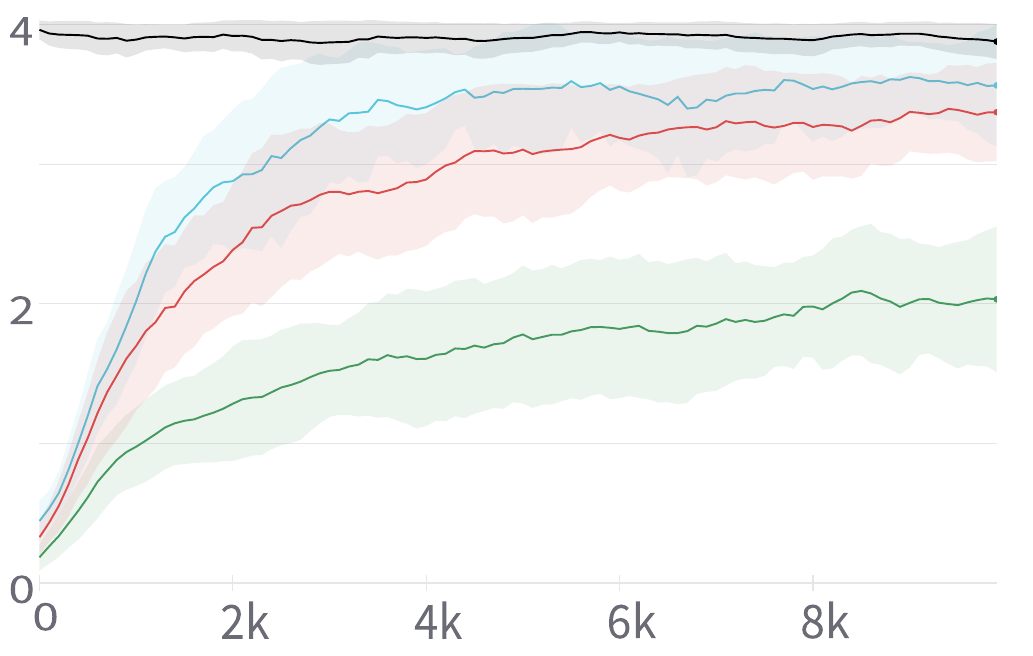}
  \caption{\pnp{2}}
\end{subfigure}
\begin{subfigure}{0.56\linewidth}
  \includegraphics[width=\linewidth]{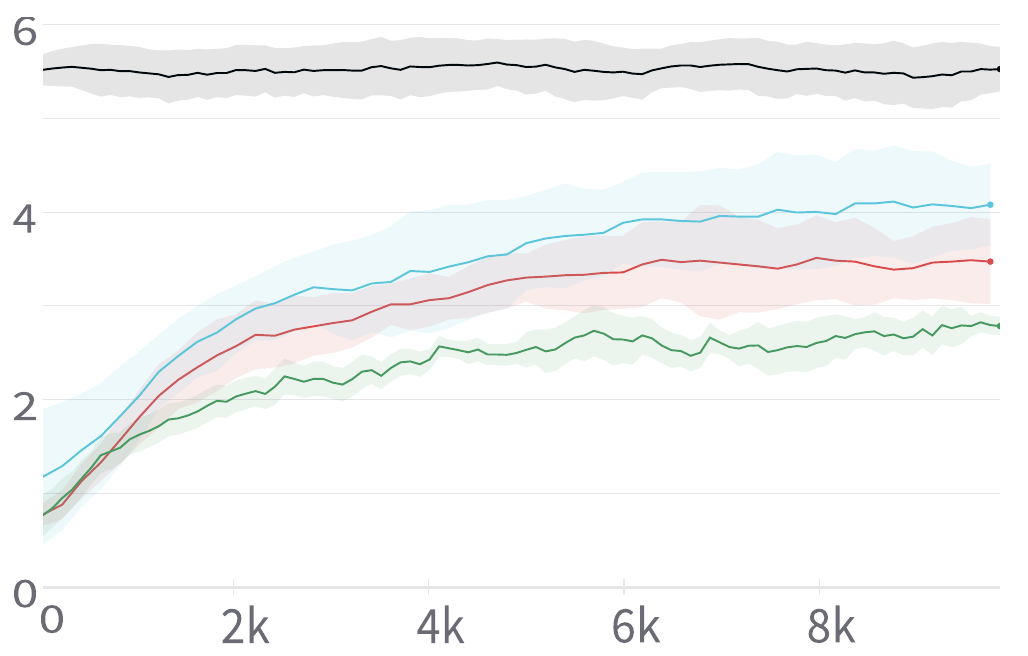}
  \caption{\pnp{3}}
\end{subfigure}%
\begin{tikzpicture}[overlay]
        \node[anchor=west] at (0.5cm,7cm) {$|D_E|:|D_I|$};
        
        \draw[mygreen,line width=1pt] (0.7cm,6.2cm) -- (1.2cm,6.2cm);
        \node[anchor=west] at (1.4cm,6.2cm) {$10:90$};

        \draw[myred,line width=1pt] (0.7cm,5.5cm) -- (1.2cm,5.5cm);
        \node[anchor=west] at (1.4cm,5.5cm) {$25:75$};

        \draw[myblue,line width=1pt] (0.7cm,4.8cm) -- (1.2cm,4.8cm);
        \node[anchor=west] at (1.4cm,4.8cm) {$50:50$};

        \node[anchor=west] at (0.5cm,3cm) {$|D_E|:|D_I|$};
        
        \draw[mygreen,line width=1pt] (0.7cm,2.2cm) -- (1.2cm,2.2cm);
        \node[anchor=west] at (1.4cm,2.2cm) {$25:125$};

        \draw[myred,line width=1pt] (0.7cm,1.5cm) -- (1.2cm,1.5cm);
        \node[anchor=west] at (1.4cm,1.5cm) {$50:100$};

        \draw[myblue,line width=1pt] (0.7cm,0.8cm) -- (1.2cm,0.8cm);
        \node[anchor=west] at (1.4cm,0.8cm) {$75:75$};
\end{tikzpicture}
\caption{\semiapproach: Effect of number of expert demos with task segment annotation on return of learned policy (denoted on $y-$axis). The $x$-axis denotes the training iteration.} 
\label{fig:Ablation_Semi}
\end{figure}
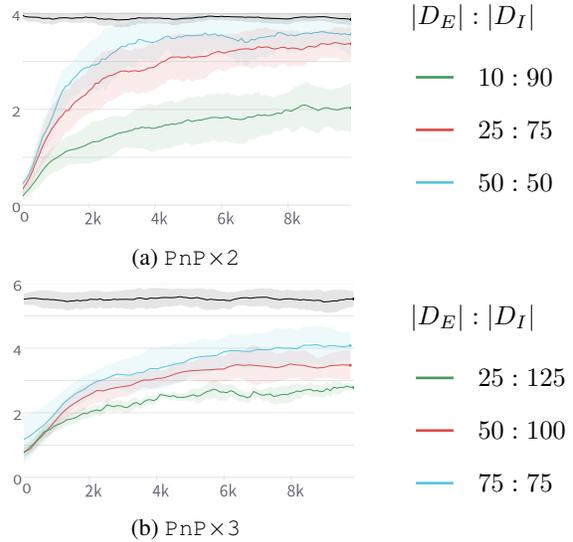

\subsection{Ablation with Semi-Supervision}
In this section, we analyze the effect of varying the proportion of expert demonstrations, $|D_E|$, to imperfect demonstrations, $|D_I|$, on \semiapproach's performance while maintaining a constant total $|D_O|$.
As illustrated in Fig.~\ref{fig:Ablation_Semi}, increasing the share of expert demonstrations boosts the convergence speed and final return for \semiapproach.
We posit that this improvement stems from the richer diversity in the expert data, ensuring broader coverage of the state-action space across task segments. 
Impressively, with just 10 annotated expert demonstrations for \pnp{2} and 25 for \pnp{3}, \semiapproach performs comparably to or outperform the unsupervised variants that utilize more expert demonstrations (see  Fig.\ref{subfig:PnP Two Object} and \ref{subfig:PnP Three Object} for reference).
\textit{This suggests that leveraging task segment annotations, which in certain applications can be cost-effective compared to obtaining additional expert demonstrations, allows \semiapproach to excel even when it has to learn from fewer expert demonstrations compared to its non-hierarchical counterparts.}

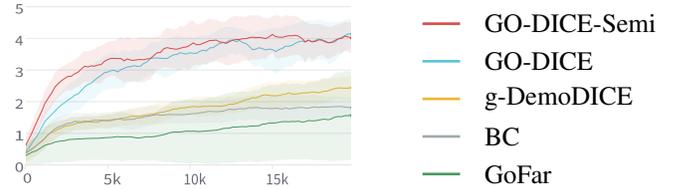
\begin{figure}[t]
\begin{subfigure}{0.56\linewidth}
  \includegraphics[width=\linewidth]{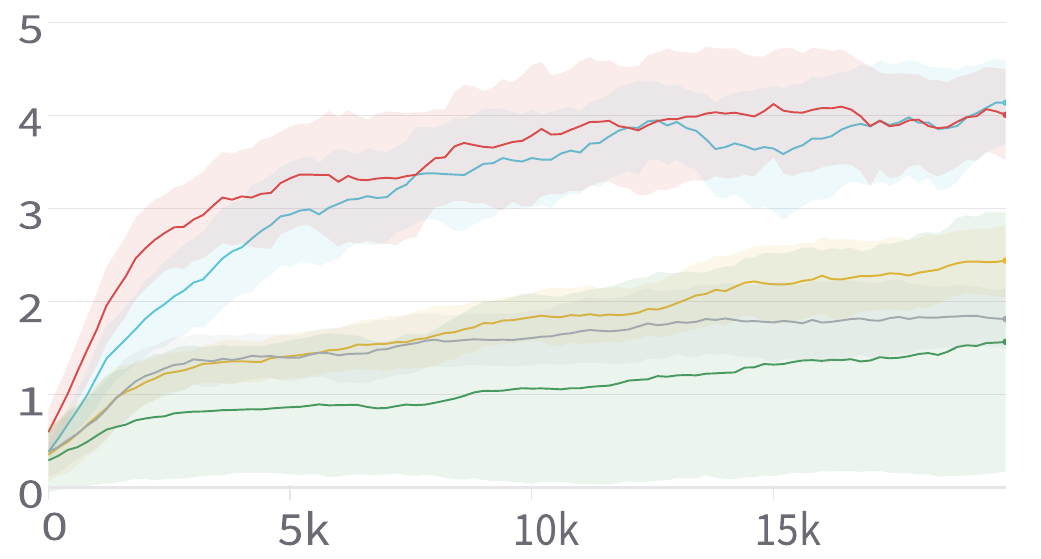}
\end{subfigure}
\begin{tikzpicture}[overlay]
        \draw[myred,line width=1pt] (0.7cm,2.2cm) -- (1.2cm,2.2cm);
        \node[anchor=west] at (1.4cm,2.2cm) {\semiapproach};
        
        \draw[myblue,line width=1pt] (0.7cm,1.7cm) -- (1.2cm,1.7cm);
        \node[anchor=west] at (1.4cm,1.7cm) {\approach};

        \draw[myyellow,line width=1pt] (0.7cm,1.2cm) -- (1.2cm,1.2cm);
        \node[anchor=west] at (1.4cm,1.2cm) {\text{g-DemoDICE}};

        \draw[mygrey,line width=1pt] (0.7cm,0.7cm) -- (1.2cm,0.7cm);
        \node[anchor=west] at (1.4cm,0.7cm) {\text{BC}};

         \draw[mygreen,line width=1pt] (0.7cm,0.2cm) -- (1.2cm,0.2cm);
        \node[anchor=west] at (1.4cm,0.2cm) {\text{GoFar}};
\end{tikzpicture}
\caption{Comparative Performance: Learning curves of \approach and the baseline techniques for \stack{3} task. $x$-axis denotes the training iteration and $y$-axis denotes the return accrued by the learned policy.}
\label{fig:PnP Stack Three Object}
\end{figure}

\subsection{Baseline Comparison on \stack{3}}
In this section, we conduct further experiments using the \stack{3} task.
This task is modeled after a common task in robotics: stacking objects.
Our implementation of the task uses the Fetch robot, Mujoco simulation environment, and three objects.
As seen in Figure~\ref{fig:PnP Stack Three Object}, our proposed \approach continues to outperform the baselines.
Interestingly, \approach with $K=3$ options mirrors the performance of \semiapproach, which used finer task segmentation annotations from expert $E_3$ with $K=9$ options.
This is contrary to the performance gap observed for \pnp{3} in Figure. \ref{subfig:PnP Three Object}.
The enhanced performance in the \texttt{Stack} environment can be attributed to the pronounced similarities across sub-tasks due to the overlapping goal space, benefiting the learning phase of \approach.
This stands in contrast to results from Figure~\ref{fig:R3} for \pnp{3} where distinct goals for each object, required finer task segmentation for improved performance.

\end{document}